\documentclass[journal]{IEEEtran}%
\usepackage[hidelinks]{hyperref}%
\usepackage[square,numbers,comma,sort&compress]{natbib}%
\usepackage{graphicx}%
\usepackage{amsmath}%
\usepackage{amssymb}%
\usepackage{subfig}%
\usepackage[table]{xcolor}%
\usepackage{trimclip}%
\newtheorem{theorem}{Theorem}%
\protected\gdef\ofel{\ensuremath{f}}%
\protected\gdef\ofeli#1{\mbox{\ensuremath{\ofel_{#1}}}}%
\protected\gdef\ofelb#1{\mbox{\ensuremath{\ofel\!\left(#1\right)}}}%
\protected\gdef\scale{\ensuremath{s}}%
\protected\gdef\upperBound{\mbox{\ensuremath{U\!B}}}%
\protected\gdef\solspel{\ensuremath{x}}%
\protected\gdef\solspeli#1{\mbox{\ensuremath{\solspel_{#1}}}}%
\protected\gdef\solspelval#1{\mbox{\ensuremath{\solspel\!\left[#1\right]}}}%
\protected\gdef\ffaH{\ensuremath{H}}%
\protected\gdef\ffaHp{\ensuremath{H'}}%
\protected\gdef\ffaHb#1{\mbox{\ensuremath{\ffaH\!\left[#1\right]}}}%
\protected\gdef\ffaHpb#1{\mbox{\ensuremath{\ffaHp\!\left[#1\right]}}}%
\protected\gdef\paramMu{\mbox{\ensuremath{\mu}}}%
\protected\gdef\paramLambda{\mbox{\ensuremath{\lambda}}}%
\protected\gdef\ea{\mbox{\textnormal{EA}}}%
\protected\gdef\fea{\mbox{\textnormal{FEA}}}%
\protected\gdef\mplea#1#2{\mbox{(#1+#2)-\ea}}%
\protected\gdef\mpleaX{\mbox{\mplea{\paramMu}{\paramLambda}}}%
\protected\gdef\mplfea#1#2{\mbox{(#1+#2)-\fea}}%
\protected\gdef\opoea{\mplea{1}{1}}%
\protected\gdef\opofea{\mplfea{1}{1}}%
\protected\gdef\opoeagz{\mbox{\ensuremath{\opoea_{>0}}}}%
\protected\gdef\opofeagz{\mbox{\ensuremath{\opofea_{>0}}}}%
\protected\gdef\countones#1{\mbox{\ensuremath{\left|#1\right|_1}}}%
\protected\gdef\onemax{\mbox{OneMax}}%
\protected\gdef\trap{\mbox{Trap}}%
\protected\gdef\leadingones{\mbox{LeadingOnes}}%
\protected\gdef\twomax{\mbox{TwoMax}}%
\protected\gdef\jump{\mbox{Jump}}%
\xdef\jumpWidthRaw{w}%
\protected\gdef\jumpWidth{\mbox{\ensuremath{\jumpWidthRaw}}}%
\protected\gdef\plateau{\mbox{Plateau}}%
\global\let\plateauWidthRaw\jumpWidthRaw%
\global\let\plateauWidth\jumpWidth%
\protected\gdef\wmodel{\mbox{W-Model}}%
\protected\gdef\maxsat{\mbox{MaxSat}}%
\protected\gdef\jssp{\mbox{JSSP}}%
\protected\gdef\runtime{\mbox{\textnormal{RT}}}%
\protected\gdef\medianRuntime{\mbox{\ensuremath{\textnormal{med}(\runtime)}}}%
\protected\gdef\meanRuntime{\mbox{\ensuremath{\textnormal{mean}(\runtime)}}}%
\protected\gdef\ert{\mbox{\textnormal{E\runtime}}}%
\protected\gdef\bigOof#1{\mbox{\ensuremath{{\mathbf O}\!\left(#1\right)}}}%
\protected\gdef\bigOmegaOf#1{\mbox{\ensuremath{{\mathbf \Omega}\!\left(#1\right)}}}%
\protected\gdef\bigThetaOf#1{\mbox{\ensuremath{{\mathbf \Theta}\!\left(#1\right)}}}%
\protected\gdef\npPrefix{\ensuremath{\mathcal{NP}}}%
\protected\gdef\npHard{\mbox{\npPrefix-hard}}%
\protected\gdef\booleans{\mbox{\ensuremath{\{\hspace{-0.1em}0,1\hspace{-0.1em}\}}}}%
\protected\gdef\runs{\ensuremath{r}}%
\protected\gdef\successFrac{\mbox{\ensuremath{\textnormal{fs}}}}%
\protected\gdef\bestSoFarF{\ofeli{\textnormal{B}}}%
\protected\gdef\meanBestF{\mbox{\ensuremath{\textnormal{mean}(\bestSoFarF)}}}%
\protected\gdef\opoeaRepCorrection{\mbox{\textcolor{violet}{\ensuremath{%
1\!-\!\left(1\!-\!\scale^{-1}\right)^{\scale}%
}}}}%
\protected\gdef\opoeaRepCorrected#1{\mbox{\ensuremath{\textcolor{violet}{\ensuremath{\left[\opoeaRepCorrection\right]}}\left[#1\right]}}}%
\protected\gdef\wmn{\ensuremath{n}}%
\protected\gdef\wmm{\ensuremath{m}}%
\protected\gdef\wmmn{\ensuremath{\wmm\wmn}}%
\protected\gdef\wmg{\ensuremath{\gamma}}%
\protected\gdef\wmnu{\ensuremath{\nu}}%
\protected\gdef\maxSatClauses{\ensuremath{c}}%
\global\let\maxSatVariables\scale%
\protected\gdef\maxSatFormula{\mbox{\ensuremath{B}}}%
\protected\gdef\maxSatFormulab#1{\mbox{\ensuremath{\maxSatFormula\!\left(#1\right)}}}%
\protected\gdef\intFromTo#1#2{\ensuremath{[#1..#2]}}%
\protected\gdef\jsspMachines{\ensuremath{M}}%
\protected\gdef\jsspJobs{\ensuremath{N}}%
\protected\gdef\jsspMA{\textnormal{MA}}%
\protected\gdef\jsspFFAMA{\mbox{\textnormal{F\jsspMA}}}%
\protected\gdef\tblbetter#1{\textbf{#1}}%
\protected\gdef\tblbettertxt{\tblbetter{bold}}%
\protected\gdef\tblsolved#1{\cellcolor{green!33!white}#1}%
\protected\gdef\tblsolvedtxt{\textcolor{green}{green~shading}}%
\let\tblsolvedReliably\tblsolved%
\protected\gdef\instance#1{\mbox{\textnormal{#1}}}%
\global\let\mapsto\rightarrow%
\begin{document}
\author{Thomas~Weise~\IEEEmembership{Member,~IEEE}, Zhize~Wu, Xinlu~Li, and Yan~Chen%
\thanks{%
Institute of Applied Optimization, %
School of Artificial Intelligence and Big Data, %
Hefei University, %
Hefei, Anhui, China~230601. %
Corresponding author: T.~Weise, tweise@hfuu.edu.cn and tweise@ustc.edu.cn%
}}%
\title{Frequency Fitness Assignment: Making Optimization Algorithms Invariant under Bijective Transformations of the Objective Function Value}%
\markboth{%
Accepted by IEEE Transactions on Evolutionary Computation%
}{%
Weise et al.,%
Frequency Fitness Assignment 2%
}%
\maketitle%
\begin{abstract}%
Under Frequency Fitness Assignment~(FFA), the fitness corresponding to an objective value is its encounter frequency in fitness assignment steps and is subject to minimization.
FFA renders optimization processes invariant under bijective transformations of the objective function value.
On TwoMax, Jump, and Trap functions of dimension~s, the classical \mbox{(1+1)-EA} with standard mutation at \mbox{rate~1/s} can have expected runtimes exponential in~s.
In our experiments, a \mbox{(1+1)-FEA}, the same algorithm but using FFA, exhibits mean runtimes that seem to scale as \mbox{s\textsuperscript{2}~ln~s}.
Since Jump and Trap are bijective transformations of OneMax, it behaves identical on all three.
On OneMax, LeadingOnes, and Plateau problems, it seems to be slower than the \mbox{(1+1)-EA} by a factor linear in~s.
The \mbox{(1+1)-FEA} performs much better than the \mbox{(1+1)-EA} on \mbox{W-Model} and MaxSat instances.
We further verify the bijection invariance by applying the \mbox{Md5~checksum} computation as transformation to some of the above problems and yield the same behaviors.
Finally, we show that FFA can improve the performance of a memetic algorithm for job shop scheduling.%
\end{abstract}%
\let\oldtheffotnote\thefootnote%
\let\thefootnote\relax\footnote{\copyright\ 2020 IEEE. Personal use of this material is permitted. Permission from IEEE must be obtained for all other uses, in any current or future media, including reprinting/republishing this material for advertising or promotional purposes, creating new collective works, for resale or redistribution to servers or lists, or reuse of any copyrighted component of this work in other works.}%
\let\thefootnote\oldtheffotnote%
\setcounter{footnote}{0}%
\section{Introduction}%
\IEEEPARstart{F}{itness} assignment is a component of many Evolutionary Algorithms (EAs).
It transforms the features of candidate solutions, such as their objective value(s), to scalar values which are then the basis for selection.
Frequency Fitness Assignment (FFA)~\cite{WWTWDY2014FFA,WWTY2014EEIAWGP} was developed to enable algorithms to escape from local optima.
In FFA, the fitness corresponding to an objective value is its encounter frequency so far in fitness assignment steps and is subject to minimization.
As we discuss in detail in Section~\ref{sec:ffa}, FFA turns a static optimization problem into a dynamic one where objective values that are often encountered will receive worse and worse fitness.

In this article, we uncover a so-far unexplored property of FFA:
It is invariant under any bijective transformation of the objective function values.
This is the strongest invariance known to us and encompasses all order-preserving mappings.
Other examples for bijective transformations include the negation, permutation, or even encryption of the objective values.
According to~\cite{OAAH2017IGOAAUPVIP}, \emph{invariance extends performance observed on a single function to an entire associated invariance class, that is, it generalizes from a single problem to a class of problems. Thus it hopefully provides better robustness w.r.t. changes in the presentation of a problem.}
FFA generalizes the performance of an algorithm on \onemax\ to all problems which are bijections thereof, including \jump\ and \trap.

While invariances are generally beneficial for optimization algorithms~\cite{W1989TGAASPWRBAORTIB,HRMSA2011IOIISWCAPFICANSP}, such strong invariance comes at a cost:
The idea that solutions of better objective values should be preferred to those with worse ones can no longer be applied, since many bijections are not order-preserving.
FFA only considers whether objective values are equal or not.
One would expect that this should lead to a loss of performance.
We find that the opposite is the case on several benchmarks evaluated in our study.
On those where FFA increases the number of function evaluations~(FEs) to find the optimum, i.e., the runtime, it seems to do so only linearly with the number of different objective values or the problem dimension~\scale, as both cases are indistinguishable in the investigated problems.

We plug FFA into the most basic EA~\cite{BFM1997HOEC}, the~\opoea\ with standard mutation at \mbox{rate~$1/\scale$}, and obtain the~\opofea.
We investigate its performance on several well-known problems, namely the \onemax, \leadingones, \twomax, \jump, \trap, and \plateau\ functions, the \wmodel, and \maxsat, all defined over bit strings of length~\scale.
We find that the resulting \opofea\ is slower on \onemax, \leadingones, and on the \plateau\ functions, while it very significantly reduces the runtime needed to solve the other problems.
Most notably, in our experiments, it has runtime requirements in the scale \mbox{of~$\scale^2\ln{\scale}$} on the \twomax, \trap\ and \jump\ problems, for which the expected runtime needed by the \opoea\ to find the global optimum is in~\bigOmegaOf{\scale^{\scale}}, \bigThetaOf{\scale^{\scale}}, and~\bigThetaOf{\scale^{\jumpWidthRaw}+\scale\ln{\scale}} (for jump width~\jumpWidth) FEs, respectively.
We confirm the invariance under bijections of the objective value by solving several benchmark problems with the \opofea\ by optimizing the \mbox{Md5~checksums}, i.e., cryptographic hashes, of their objective values and observing no change in algorithm behavior.
We also explore plugging FFA into a well-performing algorithm for a job shop scheduling problem, where it can improve the result quality under budget constraints.

In Section~\ref{sec:ffa}, we discuss the invariance property of FFA and how FFA can be plugged into the \opoea.
Related works are discussed in Section~\ref{sec:relatedWork}.
Our comprehensive experimental study is given in Section~\ref{sec:experiment}.
We conclude our article and give pointers to future work in Section~\ref{sec:conclusions}.%
\section{Frequency Fitness Assignment}%
\label{sec:ffa}%
This study investigates the impact of FFA when plugged into the maybe most basic EA, the \opoea.
The \opoea\ starts with a random bit string~\solspeli{c} of length~\scale.
Until the termination criterion is met, in each step, it applies the standard mutation operator, where each of the \scale~bits of~\solspeli{c} is flipped independently with \mbox{probability~$1/\scale$} and the result is a new string~\solspeli{n}.
If~\solspeli{n} is at least as good as~\solspeli{c}, it replaces~\solspeli{c}.
The expected runtime of the \opoea\ for an arbitrary objective function is at \mbox{most~$\scale^{\scale}$}~\cite{DJW2002OTAOTOPOEA}.
Some of the benchmark problems we investigate invoke this boundary.

We apply a slight modification of the \opoea, called the \opoeagz~\cite{CPD2017TAMPARAOEA}:
The standard mutation in each iteration is repeated until at least one bit is flipped~\cite{M1992HGARWMAH}.
No FE is wasted by evaluating a candidate solution identical to the current one.
The probability of this in the \opoea\ is~\mbox{$\left(1-\scale^{-1}\right)^{\scale}$}, which approaches~\mbox{$1/e\approx 0.368$} for~\mbox{$\scale\rightarrow\infty$}.
This small change thus saves more than one third of the FEs while not changing any other characteristic of the algorithm~\cite{CPD2017TAMPARAOEA,LO2018TAOSSA}.
In the following text, expected runtimes for the \opoea\ will therefore be corrected by factor~\opoeaRepCorrection\ to hold for the \opoeagz\ where necessary.%
\begin{figure}%
\setlength{\fboxsep}{0pt}%
\subfloat[\opoeagz]{%
\includegraphics{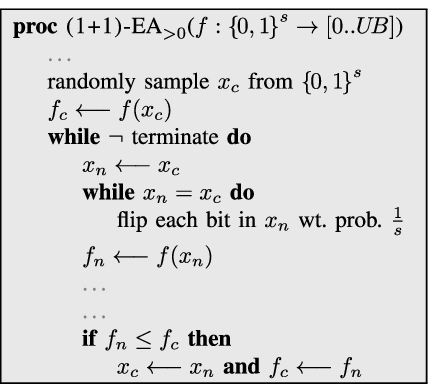}%
\label{fig:impl:ea}%
}%
\hfill%
\subfloat[\opofeagz]{%
\includegraphics{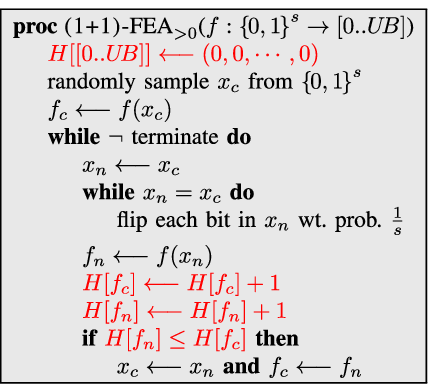}%
\label{fig:impl:fea}%
}%
\caption{The simplified pseudo codes of the \opoeagz\ and the \opofeagz, which applies FFA, for minimization problems. %
Differences are marked in \textcolor{red}{red}. %
Note: In an actual implementation, the algorithms would remember and return the candidate solution with the best encountered \emph{objective value}~\bestSoFarF\ (not fitness).}%
\label{fig:eaAndFea}%
\end{figure}%

In Figure~\ref{fig:eaAndFea}, we put the pseudo code of the \opoeagz\ next to a simplified version of the \opofeagz.
We assume that%
\begin{enumerate}%
\item the objective function~\ofel\ is subject to minimization, that%
\item its upper bound~\upperBound\ is known, that%
\item all objective values are integers greater or equal to~0, and that%
\item the solution space is~\mbox{$\booleans^{\scale}$}, the bit strings of length~\scale.%
\end{enumerate}%
This can be established for many well-known benchmark problems on which the \opoea\ is usually investigated, as well as for many practical optimization problems like \maxsat.

Under these assumptions, only minimal changes to the \opoeagz\ are necessary to introduce FFA:
An array~\ffaH\ of integers of length~\mbox{$\upperBound+1$} is used to hold the frequency of each objective value in~\mbox{$\intFromTo{0}{\upperBound}$}.
Before selecting one of the two candidate solutions with objective values~\ofeli{c} and~\ofeli{n}, the frequencies~\ffaHb{\ofeli{c}} \emph{and}~\ffaHb{\ofeli{n}} of these objective values are increased.
The results of these increments are compared.
Note: Both frequencies are increased, because if \ffaHb{\ofeli{c}} was not incremented, solutions with unique objective values could become traps for the optimization process.

In order to conduct an efficient search under FFA, the set~${\mathbb{Y}}$ of possible objective values for the problem to be solved should not be too big.
FFA must maintain a frequency table~\ffaH, which has the same size as~$\mathbb{Y}$.
Also, FFA attempts to distribute the search effort evenly over all objective values.
In the extreme case where each distinct solution has a different objective value, FFA almost degenerates the search to a random walk.

Most often, the \opoea\ is analyzed as maximization algorithm.
Since the \opofea\ minimizes the objective value frequencies, we also present the \opoea\ for minimization and define the benchmark problems in Section~\ref{sec:experiment} accordingly.

The \opofea\ implementation given in Figure~\ref{fig:impl:fea} can easily be extended towards a \mpleaX.
It can also be modified to handle problems with unknown upper and lower bounds of the objective function (or objective functions that return real numbers but can still be discretized) by implementing~\ffaH\ as hash table~\cite{CLRS2009HT} (see Section~\ref{sec:md5}).
FFA can be introduced into arbitrary metaheuristics.%
\begin{theorem}%
\label{theo:invariance}%
The sequence of candidate \mbox{solutions~$\solspel\in\mathbb{X}$} generated by an optimization process applying FFA is invariant under any \mbox{bijection~$g:\mathbb{Y}\mapsto\mathbb{Z}$} of the objective \mbox{function~$\ofel:\mathbb{X}\mapsto\mathbb{Y}$}, where \mbox{$\mathbb{X}$~is} the solution space, \mbox{$\mathbb{Y}$~is} a finite subset of~$\mathbb{R}$, \mbox{and~$\mathbb{Z}$} is a set of the same size.%
\begin{IEEEproof}%
The bijection~$g$ maps each value from~$\mathbb{Y}$ to one value in~$\mathbb{Z}$ and vice versa.
Therefore, if two objective values identify the same (or a different) entry in~\ffaH, so will their bijective transformations.
Under FFA, \emph{only} the entries in~\ffaH\ are modified and compared to make selection decisions.%
\end{IEEEproof}%
\end{theorem}

We can also prove this inductively:
Assume that two runs of the \opofeagz\ which minimize \ofel\ and $g\circ\ofel$, respectively, are identical until iteration~$t$:
They have the same random seed, same~$\solspeli{c}$, and $\ffaHb{y}=\ffaHpb{g(y)}\forall y\in\mathbb{Y}$ holds for their respective FFA tables.
Both will sample the same next point~$\solspeli{n}$.
$\ffaHb{\ofelb{\solspeli{c}}}=\ffaHpb{g(\ofelb{\solspeli{c}})}$ and $\ffaHb{\ofelb{\solspeli{n}}}=\ffaHpb{g(\ofelb{\solspeli{n}})}$ will still hold after incrementing the entries.
Hence, both will make the same decision regarding the update of~\solspeli{c} and begin iteration $t+1$ in the same state.
In Sections~\ref{sec:jump}, \ref{sec:trap}, and~\ref{sec:md5}, we provide experimental evidence that this invariance indeed holds.%
\section{Related Work}%
\label{sec:relatedWork}%
FFA was designed as an approach to prevent the premature convergence to a local optimum.
In the context of EAs, it is therefore related to fitness sharing, niching, and clearing~\cite{GR1987GAS,P1996ACPAANMFGA}.
Several such diversity-preserving mechanisms have been plugged into a \mplea{\paramMu}{1} and studied theoretically in~\cite{FOSW2009AODPMFGE} on \twomax, where the original algorithm requires~\bigOof{\scale^\scale}~FEs to find the optimum.
It is found that avoiding fitness or genotype duplicates does not help, whereas with deterministic crowding and sufficiently large~$\paramMu$, the problem can be solved efficiently with high probability.
With fitness sharing and $\paramMu\geq2$, the \mplea{\paramMu}{1} can solve \twomax\ in~\bigOof{\paramMu\scale\log{\scale}}.
Different from the above methods, which only consider the current population, FFA tries to guide the search away from objective values that have been encountered often during the whole course of the optimization process.
In Section~\ref{sec:twomax}, we will show that FFA can help solving \twomax\ efficiently already at~$\paramMu=1$.

Another related idea is Tabu Search (TS)~\cite{GTDW1993AUGTTS}, which improves local search by declaring solutions (or solution traits) which have already been visited as tabu, preventing them from being sampled again.
Like FFA, it utilizes the search history, but usually in form of a list of tabu solutions or solution traits.
Different from FFA, the TS relies on the order of objective values when deciding which solutions to accept.

The Fitness Uniform Selection Scheme (FUSS)~\cite{H2002FUSTPGD} selects solutions in such a way that their corresponding objective values are approximately uniformly distributed within the range of the minimum and maximum objective value in the population.
The Fitness Uniform Deletion Scheme (FUDS)~\cite{HL2006FUO} works similarly, but instead of selecting individuals, it deletes them when slots in the population are required to integrate the offspring.
Both methods need populations, only consider the individuals in the current population, and are only invariant under translation and scaling of the objective function.

The ageing operator in Artificial Immune Systems (AIS) deletes individuals either after they have survived a certain number of iterations, with a certain probability, or both~\cite{COY2020WHAAEAISTOEA,COY2018FAIS,COY2019AISCFAGAFTNNPP}.
Ageing has also been applied in EAs~\cite{GTT1996IAIGA}.
Like FFA, ageing makes solutions less attractive if they remain in the population for a long time.
Different from FFA, the information about these solutions disappears from the optimization process once they ``die.''

Methods which try to balance between solution quality and population diversity are today grouped under the term Quality-Diversity (QD) algorithms~\cite{GLY2019QDTS,CD2018QADOAUMF,PSS2016QDANFFEC}.

Novelty Search (NS)~\cite{LS2011AOETTSFNA} is a QD algorithm.
Instead of an objective function~\ofel, NS uses a (dynamic) novelty metric~$\rho$.
This metric is computed, e.g., as mean behavior difference to the $k$~nearest neighbors in an archive of past solutions.
FFA works on the original objective function and just transforms it to a dynamic fitness measure.
It does not require an archive of solutions but uses a table~\ffaH\ counting the frequency of the objective values.

While NS was aimed to abandon the objective function~\ofel, using it as behavior definition was also tested~\cite{LS2011AOETTSFNA}.
Then, $\rho$~is the mean distance to $k$~neighbors (or all solutions ever found) in the objective space.
Unlike FFA, this uses the assumption that differences between objective values are useful or correlate with diversity.
Novelty Search with Local Competition (NSLC)~\cite{LS2011EADOVCTNSALC} combines the search for finding diverse solutions with a local competition objective rewarding solutions which can outperform those most similar to them.

The \mbox{MAP-Elites} algorithm~\cite{MC2015ISSBME} combines a performance objective~\ofel\ and a user-defined space of features that describe candidate solutions, which is not required by FFA.
\mbox{MAP-Elites} searches for highest-performing solution in each cell of the discretized feature space.

Surprise Search (SS)~\cite{GLY2016SSBOAN} uses the concept of surprise as an alternative to novelty.
A solution is scored by the difference of its observed behavior from the expected behavior.
A history of discovered solution behaviors is maintained and used to predict the behavior of the new solutions.
SS has also been combined with NSLC in a multi-objective fashion~\cite{GLY2019QDTS}.

All of the above algorithms are conceptually different from FFA.
They either are complete optimization methods (NS, QD, TS) or modules for EAs (FUSS/FUDS), while FFA can be plugged into many different optimization algorithms.
Unlike FFA, none of the above methods exhibits an invariance under bijective transformations of the metrics they try to optimize.

From the perspective of invariances, FFA is related to Information-Geometric Optimization (IGO)~\cite{OAAH2017IGOAAUPVIP}.
IGO also replaces the objective function~\ofel\ with an adaptive transformation of it.
This transformation indicates how good or bad an objective value is relative to other observed objective values, i.e., is different from our method which simply compares encounter frequencies.
IGO is invariant under all strictly increasing transformations of~\ofel, whereas FFA creates invariance under all bijective transformations.
IGO is a complete family of optimization methods which can also exhibit invariance under several transformations of the search space.
Since FFA only works on~\ofel, it cannot provide such invariances.
IGO can optimize continuous objective functions, which is not possible with FFA.%
\section{Experiments}%
\label{sec:experiment}%
We now apply the \opoeagz\ and the \opofeagz\ to minimization versions of different classical optimization problems.
We initialize the \opoeagz\ and the \opofeagz\ with the same random seeds for each run, i.e., we always have pairs of runs starting at the same random initial solution and sampling the same first offspring solutions for both algorithms.

The runs are terminated when they discover the optimum.
In some experiments, we additionally limit the computational budget to \mbox{$10^{10}=10'000'000'000$~FEs}.
This should be enough to converge on problems that the algorithms can solve well, as can be seen in Section~\ref{sec:leadingones}.
Leading to several hours to more than a day for a single run on the corresponding problems, this was also the maximum budget we could feasibly allow.

Whenever all runs on an instance succeed to find the optimum, we can compare the mean runtime `\meanRuntime' they need to do so in terms of the consumed~FEs (often called the \emph{first hitting time}).
When some runs fail in the budget-limited settings, we follow the approach from~\cite{HAFR2012RPBBOBES} and use the empirically estimated expected runtime (\ert) instead.
The \ert\ for a problem instance is estimated as the ratio of the sum of all FEs that all the runs consumed until they \emph{either} have discovered a global optimum \emph{or} exhausted their budget, divided by the number of runs that discovered a global optimum~\cite{HAFR2012RPBBOBES}.
The \ert\ is the mean expected runtime under the assumption of independent restarts after failed runs, which then may either succeed (consuming the mean runtime of the successful runs) or fail again (with the observed failure probability, after consuming ${10}^{10}$~FEs).

In order to guarantee the reproducibility of our work, we provide the complete data used in this paper, including the result log files, the scripts used to generate all the figures and tables, as well as the source code of all algorithms and all benchmark problems in~\cite{WWLC2019TEDFTSFFAMOAIUBTOTOF}.%
\subsection{\onemax~Problems}%
\begin{figure}[tb]%
\centering%
\includegraphics[width=0.9999\linewidth]{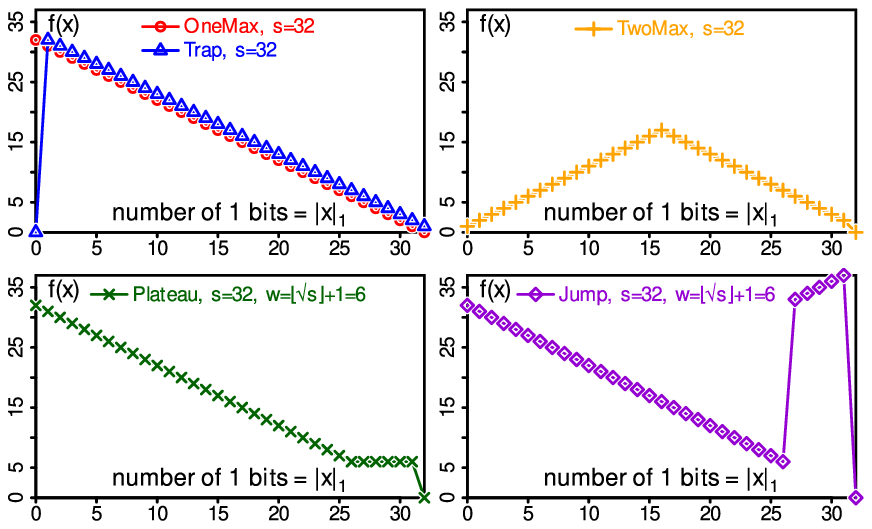}
\caption{Illustrations of the \onemax, \twomax, \trap, \jump, and \plateau\ problems for $\scale=32$ and $\jumpWidth=6$.}%
\label{fig:function_illustrations}%
\end{figure}%
\onemax~\cite{M1992HGARWMAH} is a unimodal optimization problem where the goal is to discover a bit string of all ones.
Its minimization version of dimension~\scale\ is defined below and illustrated in Figure~\ref{fig:function_illustrations}:%
\begin{equation}%
\onemax(\solspel) = \scale - \countones{\solspel}\textnormal{~where~}\countones{\solspel}=\sum_{i=1}^{\scale} \solspelval{i}%
\end{equation}%
It has a black-box complexity of $\bigOmegaOf{\scale/\ln{\scale}}$~\cite{ER1963OTPOIT,DJW2006UALBFRSHIBBO}.
Here, an \opoea\ has an expected runtime of $\bigThetaOf{\scale\ln{\scale}}$~FEs~\cite{M1992HGARWMAH}.
A very exact formula~\cite{HPRTC2018PAOTOPOEA} with our correction factor for the \opoeagz\ is given in Equation~\eqref{eq:opoea:onemax}, where $C_1\approx1.89254$ and $C_2\approx0.59789875$.%
\begin{equation}%
\resizebox{0.905\linewidth}{!}{\ensuremath{\opoeaRepCorrected{e\scale\ln{\scale}-C_1\scale+0.5e\ln{\scale}+C_2+\bigOof{(\ln{\scale})/\scale}}}}%
\label{eq:opoea:onemax}%
\end{equation}%
\begin{figure}[tb]%
\centering%
\includegraphics[width=0.9999\linewidth]{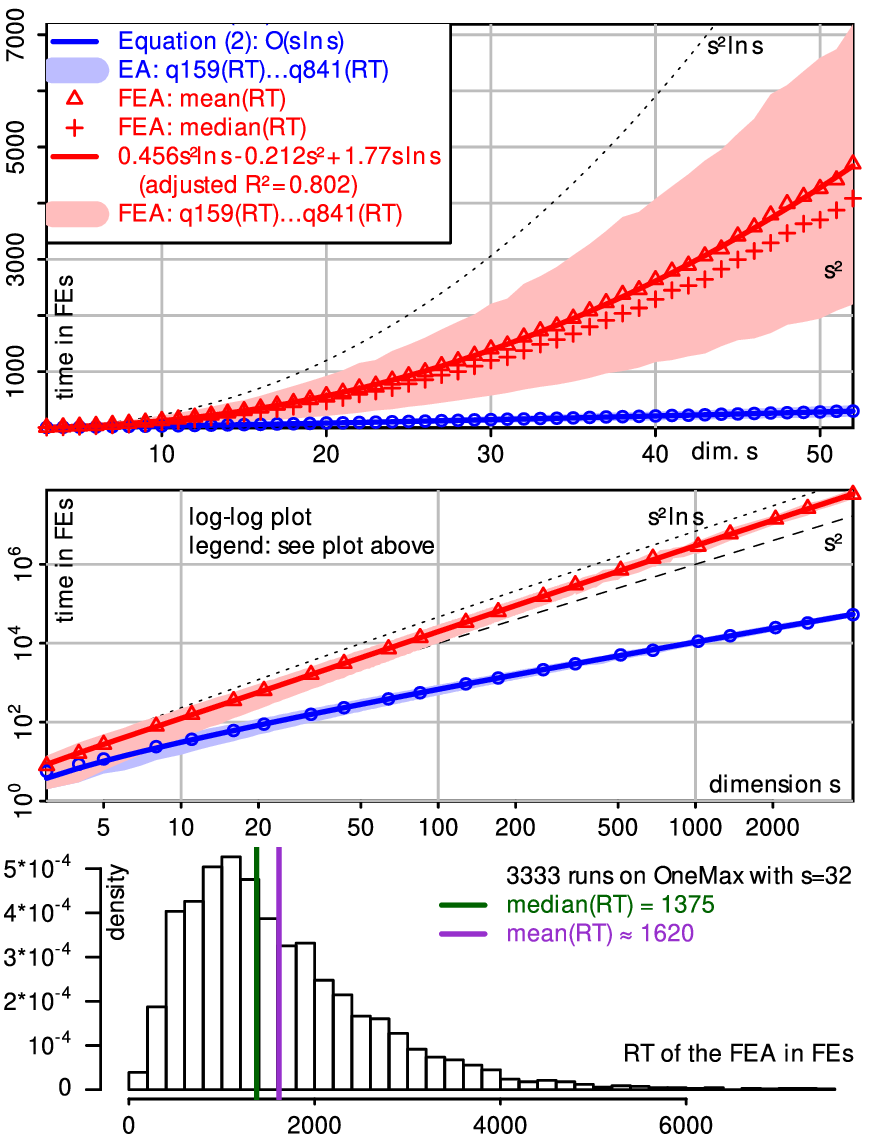}%
\caption{The runtime measured for the \opoeagz\ and \opofeagz\ on selected instances of the \onemax\ problem.}%
\label{fig:onemax_runtime}%
\end{figure}%
\begin{figure}[tb]%
\centering%
\includegraphics[width=0.9999\linewidth]{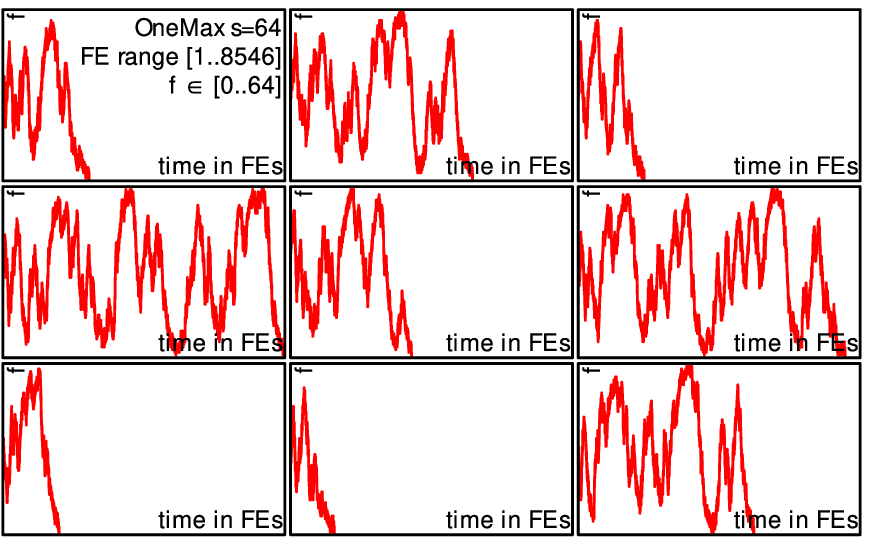}\
\caption{9~typical runs of \opofeagz\ on \onemax\ ($\scale=64$)}%
\label{fig:onemax_progress}%
\end{figure}%

We conduct 3333~runs with both the \opoeagz\ and \opofeagz\ on this problem for each~$\scale\in\intFromTo{3}{333}$ and 71~runs for 26~selected larger values of~\scale\ up to~4096, all without budget constraint.
In Figure~\ref{fig:onemax_runtime}, we illustrate the mean runtime to solve the instances with the range of the 15.9\% to the 84.1\% quantiles in the background.\footnote{%
These quantiles are wider than the inter-quartile range and would represent exactly the range mean-stddev to mean+stddev under a normal distribution.%
}
In the top-most sub-figure, we illustrate all results for $\scale\in\intFromTo{3}{52}$.
The middle figure is a log-log plot based on the complete data, but with marks only placed at $\scale\in \{2^i, round(2^i/3)\}$ to not clutter the plot.
In both diagrams, we illustrate the results of Equation~\eqref{eq:opoea:onemax} without the \mbox{\bigOof{(\ln{\scale})/\scale}~term}.
They exactly match the results of the \opoeagz.

The mean runtime of the \opofeagz\ seems to be in the scale \mbox{of~$\scale^2\ln{\scale}$} for the investigated range of~\scale.
The illustrated model was obtained using linear regression on the complete set of 1'105'069~runs with the inverse variances of the measured runtimes per distinct \scale~value used as weights.
All regression models in the rest of this article are obtained in the same way.
The curve of the model visually fits to the mean runtimes and the adjusted $R^2$~value of~$0.8$ indicates that it can explain most of the variance in the data.

The observed distribution of the runtime is skewed and the median is lower than the mean on all dimensions.
This is illustrated exemplarily in the histogram for dimension~$\scale=32$ in the lower part of Figure~\ref{fig:onemax_runtime}.
Its shape resembles a log-normal distribution or a sum of parameterized geometric distributions~\cite{D2019ARSHVSD}.
For~$s\leq8$, the histograms look like exponential distributions, caused by the high chance of randomly guessing the optimum.

Figure~\ref{fig:onemax_progress} illustrates nine typical runs of the \opofeagz\ on the \onemax\ problem with~$\scale=64$.
Initially, some of the runs progress towards better solutions, others to worse.
They change the search direction from time to time.
This oscillation is repeated until the global optimum is discovered.%
\subsection{\leadingones~Problems}%
\label{sec:leadingones}%
The \leadingones\ problem~\cite{W1989TGAASPWRBAORTIB,R1997CPOEA} maximizes the length of a leading sequence containing only 1~bits.
Its minimization version of dimension~\scale\ is defined as follows:%
\begin{equation}%
\leadingones(\solspel) = \scale - \sum_{i=1}^{\scale} \prod_{j=1}^i \solspelval{j}%
\end{equation}%
The problem exhibits epistasis, as the bit at index~2 can only contribute to the objective value if the bit at index~1 has value~1.
The black-box complexity of \leadingones\ is~$\bigThetaOf{\scale\ln{\ln{\scale}}}$~\cite{AADLMW2013TQCOFAHP}.
The \opoea\ has a quadratic expected runtime on \leadingones~\cite{DJW2002OTAOTOPOEA}.
The exact formula~\cite{BDN2010OFAAMRFTLP,S2013ANMFLBOTRTOEA} is presented with our correction factor in Equation~\eqref{eq:opoea:lo}:%
\begin{equation}%
\opoeaRepCorrected{0.5\scale^2\left( \left(1-1/\scale\right)^{1-\scale} -1 + 1/\scale\right)}%
\label{eq:opoea:lo}%
\end{equation}%
\begin{figure}[tb]%
\centering%
\includegraphics[width=0.9999\linewidth]{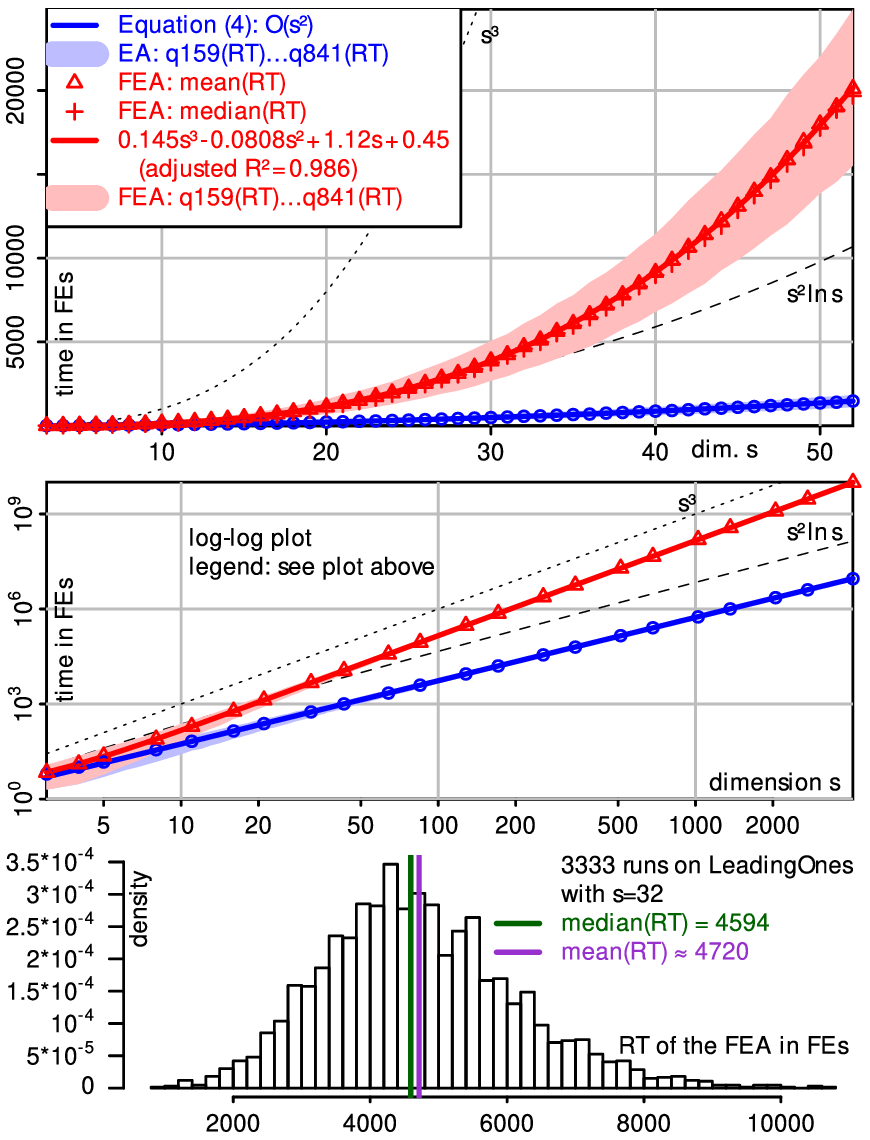}
\caption{The runtime measured for the \opoeagz\ and \opofeagz\ on the \leadingones\ problem.}%
\label{fig:leadingones_runtime}%
\end{figure}%
Figure~\ref{fig:leadingones_runtime} has the same structure as Figure~\ref{fig:onemax_runtime} and is based on an experiment with the same parameters, only using the \leadingones\ instead of the \onemax\ problem.
The \opoeagz\ behaves as predicted in Equation~\eqref{eq:opoea:lo}.

The runtime of the \opofeagz\ fits to the illustrated regression model for the investigated range of~\scale\ and can explain almost all of the variance of the data.
Due to the approximately cubic runtime, the mean time to solve \leadingones\ at $\scale=4096$ is $9.9\!\cdot\!10^{9}$~FEs.
The histogram of the observed runtimes for \mbox{dimension~$\scale=32$} in the lower part of Figure~\ref{fig:leadingones_runtime} is slightly skewed.%
\subsection{\twomax~Problems}%
\label{sec:twomax}%
The minimization version of the \twomax~\cite{FQW2018ELDBOAWHTMO,VHGN2002FTTIMEAHSP} problem of dimension~\scale\ can be defined as follows:%
\begin{equation}%
\resizebox{0.905\linewidth}{!}{\ensuremath{%
\twomax(\solspel) = \left\{\!\!\!%
\begin{array}{l@{~}l}%
0&\textnormal{if }\countones{\solspel} = \scale\\%
1+\scale-\max\{\countones{\solspel}, \scale-\countones{\solspel}\}&\textnormal{otherwise}%
\end{array}%
\right.%
}}%
\end{equation}%
\begin{figure}[tb]%
\centering%
\includegraphics[width=0.9999\linewidth]{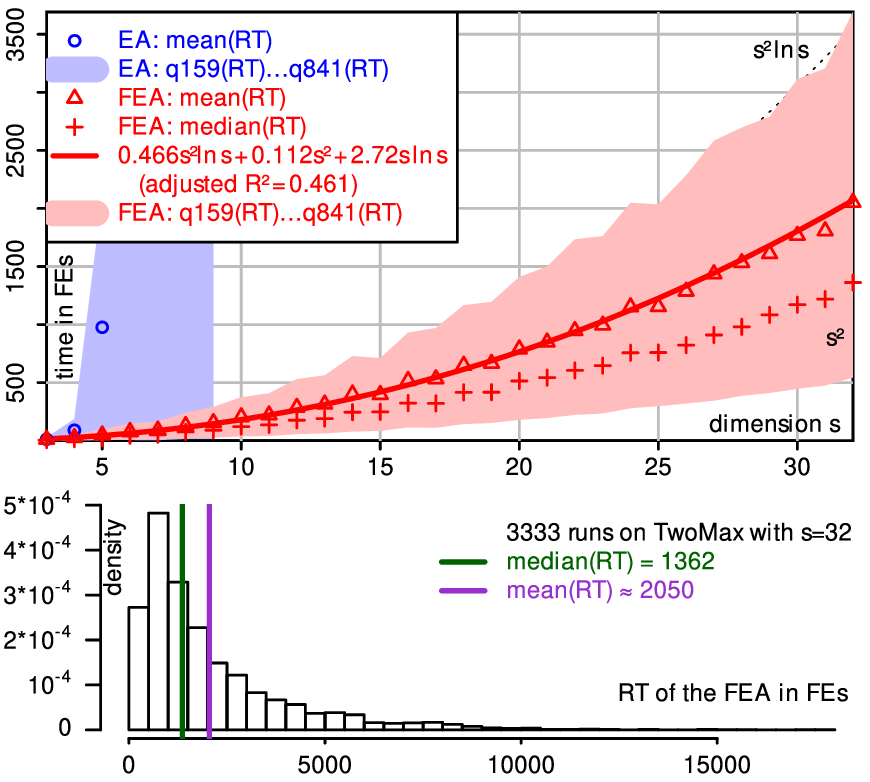}%
\caption{The runtime measured for the \opoeagz\ and \opofeagz\ on the \twomax\ problem.}%
\label{fig:twomax_runtime}%
\end{figure}%
The \twomax\ problem introduces deceptiveness in the objective function by having a local and a global optimum.
Since their basins of attraction have the same size, a \opoea\ can solve the problem in \bigThetaOf{\scale\ln{\scale}} steps with probability~0.5 while otherwise needing exponential runtime.
The resulting overall expected runtime is in \bigOmegaOf{\scale^{\scale}}~\cite{FOSW2009AODPMFGE,FQW2018ELDBOAWHTMO}.

On each instance of \twomax\ with $\scale\in\intFromTo{3}{32}$, we conduct 71~runs with \opoeagz\ and 3333~with \opofeagz.
For all experiments from here on except in Section~\ref{sec:jssp}, we use a budget of ${10}^{10}$~FEs.
The \opoeagz\ succeeds in solving the problem only in about half of the runs for~$\scale>10$ within the budget, which was the reason for the \mbox{71-run} limit.
We illustrate its performance in Figure~\ref{fig:twomax_runtime} only for dimensions~$\scale<10$ where it always succeeded.

All runs of \opofeagz\ solved their corresponding instances.
The algorithm exhibits a mean runtime fitting to a model of scale~$\scale^2\ln{\scale}$ for~$\scale\in\intFromTo{3}{32}$, which is a big improvement in comparison to the exponential time needed by the \opoea.
The lower adjusted~$R^2$ and less smooth increase of the runtime with~\scale\ result from the slightly different shapes of the \twomax\ problem for odd and even values of~\scale.
The median runtime is again smaller than the mean.
The histogram of the observed runtimes for $\scale=32$ in the lower part of Figure~\ref{fig:twomax_runtime} again exhibits the familiar skew.

These results are interesting, since avoiding fitness duplicates in a \mplea{\paramMu}{1} does not help to solve the problem efficiently~\cite{FOSW2009AODPMFGE}.
FFA thus does more than this even \mbox{at~$\paramMu=1$}.%
\subsection{\jump~Problems}%
\label{sec:jump}%
The \jump\ functions as defined in~\cite{DJW2002OTAOTOPOEA,FQW2018ELDBOAWHTMO} introduce a deceptive region of width~\jumpWidth\ with very bad objective values right before the global optimum.
The minimization version of the \jump\ function of dimension~\scale\ and jump width~\jumpWidth\ is defined as follows:%
\footnote{%
Researchers have formulated different types of \jump\ functions. %
The one in~\cite{DDK2015UBBCOJF}, e.g., is similar to our \plateau\ function but differs in the plateau objective value.%
}%
\begin{equation}%
\jump(\solspel) = \left\{\!\!\!%
\begin{array}{l@{~}l}%
\scale-\countones{\solspel}&\textnormal{if }\left(\countones{\solspel} = \scale\right)\lor\left(\countones{\solspel} \leq \scale-\jumpWidth\right)\\%
\jumpWidth+\countones{\solspel}&\textnormal{otherwise}%
\end{array}%
\right.%
\label{eq:jump}%
\end{equation}
The expected runtime of the \opoea\ on such problems is in \bigThetaOf{\scale^{\jumpWidthRaw}+\scale\ln{\scale}}~\cite{DJW2002OTAOTOPOEA}.
The \jump\ problem is a bijective transformation of the \onemax\ problem.\footnote{%
For~$\jumpWidth=1$, the \jump\ and \plateau\ problems are \onemax\ problems, which is why we do not perform or illustrate any runs with $\jumpWidth=1$ for either.%
} %
The \opofeagz\ will exhibit the same behavior and runtime requirement on \emph{any} jump problem instance as on a \onemax\ instance of the same dimension~\scale, regardless of the jump width~\jumpWidth.%
\begin{figure}[tb]%
\centering%
\includegraphics[width=0.9999\linewidth]{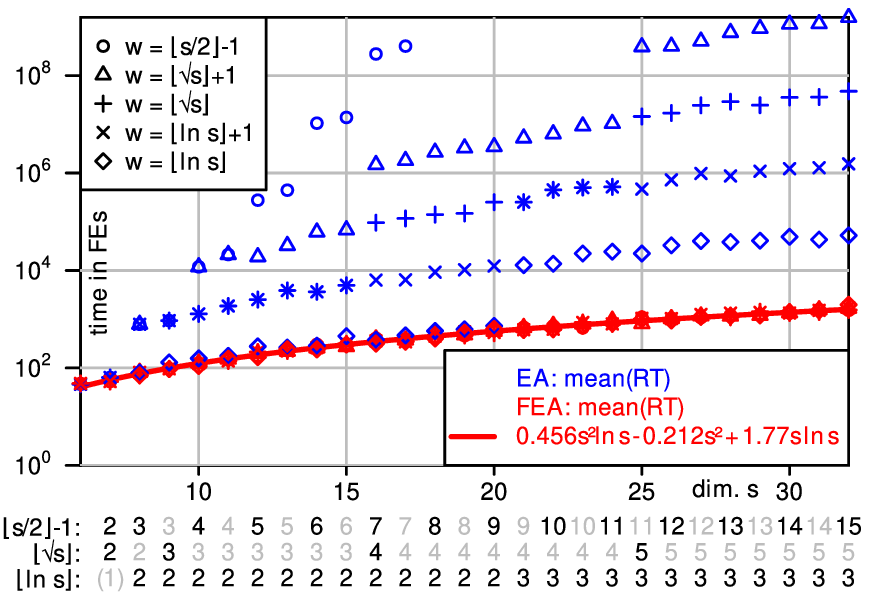}%
\caption{The runtime measured for the \opoeagz\ and \opofeagz\ on \jump\ problems with dimension~\scale\ and jump width~$\jumpWidth>1$.}%
\label{fig:jump_runtime}%
\end{figure}%

We conduct experiments with five different jump widths~\jumpWidth, namely %
\mbox{$\left\lfloor\ln{\scale}\right\rfloor$}, %
\mbox{$\left\lfloor\ln{\scale}\right\rfloor+1$}, %
\mbox{$\left\lfloor\sqrt{\scale}\right\rfloor$}, %
\mbox{$\left\lfloor\sqrt{\scale}\right\rfloor+1$}, and %
\mbox{$\left\lfloor0.5\scale\right\rfloor-1$}. %
We illustrate the results in Figure~\ref{fig:jump_runtime} only for those setups where a success rate of 100\% within the $10^{10}$~FEs were achieved in 71~runs.
\opofeagz\ finds the optimum in all runs and all the observed mean runtimes fall on the function fitted to the results on \onemax\ (see Figure~\ref{fig:onemax_runtime}), confirming that the two problems are indeed identical from the perspective of an algorithm using FFA.
As expected, the runtime of the \opoeagz\ steeply increases with the jump width~\jumpWidth\ and it is outperformed by the \opofeagz.

Depending on its configuration, the AIS \mbox{Opt-IA}~\cite{COY2020WHAAEAISTOEA} needs runtimes of at least \bigOof{\scale^2 \ln{\scale}} and \bigOof{\scale^3}~FEs on \onemax\ and \leadingones, respectively.
It seems that our \opofea\ has similar requirements, i.e., compared to the \opoea\ with standard bit mutation, a linear slowdown is incurred on these problems.
However, on the \jump\ problems, \mbox{Opt-IA} needs, again depending on its configuration, at least~\bigOof{\frac{\scale^{\jumpWidthRaw+1}\cdot e^{\jumpWidthRaw}}{\jumpWidthRaw^{\jumpWidthRaw}}}~FEs.
The \mbox{(1+1) Fast-IA}~\cite{COY2018FAIS}, which is at least as fast as the \opoea\ on \onemax\ and \leadingones, needs exponential expected runtime for sufficiently large~\jumpWidth.
The Fast~EA~\cite{DLMN2017FGA} using a heavy-tailed mutation rate, too, needs exponential runtime to solve \jump.
The asymptotic performance of the cGA on \jump\ is not worse than on \onemax\ for logarithmic jump widths~\jumpWidth~\cite{D2019ATRAFTCOJFECCFVANEC}, but it still needs exponential time for larger~\jumpWidth~\cite{D2019AELBFTROTCOJF}.
\opofea, however, performs on \jump\ exactly as on \onemax, regardless of~\jumpWidth.%
\subsection{\trap\ Function}%
\label{sec:trap}%
The \trap\ function~\cite{NB2003AAOTBOSEAOTF,DJW2002OTAOTOPOEA} is very similar to the \onemax\ problem, except that it replaces the worst possible solution there with the global optimum.
Following a path of improving objective values will \emph{always} lead the optimization algorithm away from the global optimum.
The \opoea\ here has an expected runtime of~\bigThetaOf{\scale^{\scale}}~\cite{DJW2002OTAOTOPOEA}.
The minimization version of the \trap\ function can be specified as follows:%
\begin{equation}%
\trap(\solspel) = \left\{\!\!\!%
\begin{array}{l@{~}l}%
0&\textnormal{if~}\countones{\solspel}=0\\%
\scale-\countones{\solspel}+1&\textnormal{otherwise}%
\end{array}\right.%
\end{equation}%
\begin{figure}[tb]%
\centering%
\includegraphics[width=0.9999\linewidth]{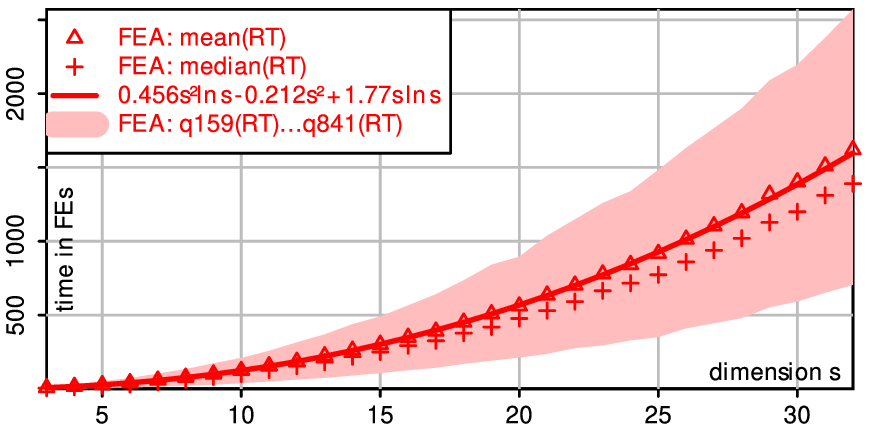}%
\caption{The runtime measured for the \opofeagz\ on the \trap\ problem.}%
\label{fig:trap_runtime}%
\end{figure}%
The \trap\ function is another bijective transformation of the \onemax\ problem.
When we plot the results from 3333~runs of the \opofeagz\ on the \trap\ function in Figure~\ref{fig:trap_runtime}, we find that the results are almost exactly identical to those obtained on \onemax\ and illustrated in Figure~\ref{fig:onemax_runtime}.
The function fitted to the mean runtime on \onemax, again plotted in Figure~\ref{fig:trap_runtime}, passes through the points measured on the \trap\ function.%
\subsection{\plateau~Problems}%
The minimization version of the \plateau~\cite{AD2018PRAFP} function of dimension~\scale\ with plateau width~\plateauWidth\ is defined as follows:%
\begin{equation}%
\resizebox{0.905\linewidth}{!}{\ensuremath{%
\plateau(\solspel) = \left\{\!\!\!%
\begin{array}{l@{~}l}%
\scale-\countones{\solspel}&\textnormal{if }\left(\countones{\solspel} = \scale\right)\lor\left(\countones{\solspel} \leq \scale-\plateauWidth\right)\\%
\plateauWidth&\textnormal{otherwise}%
\end{array}%
\right.%
}}%
\end{equation}%
\begin{figure}[tb]%
\centering%
\includegraphics[width=0.9999\linewidth]{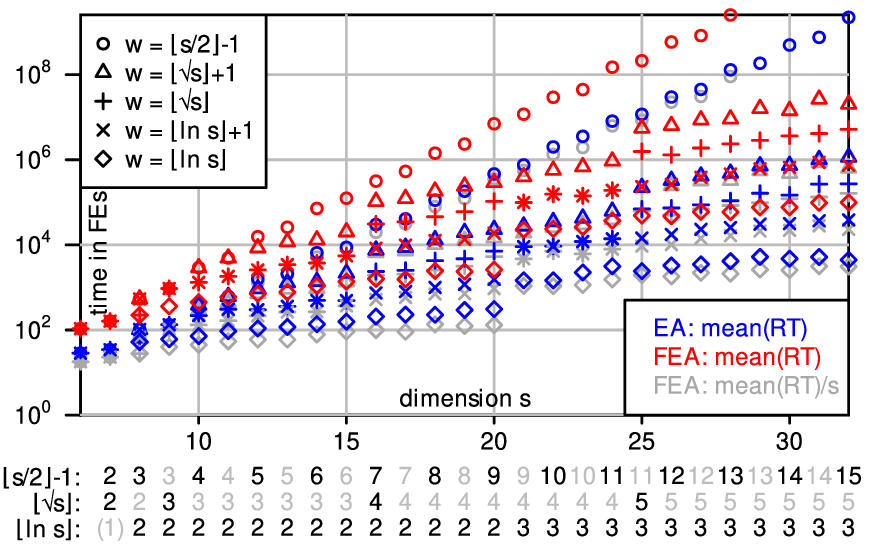}%
\caption{The runtime measured for the \opoeagz\ and \opofeagz\ on \plateau\ problems with dimension~\scale\ and plateau width~$\plateauWidth>1$.}%
\label{fig:plateau_runtime}%
\end{figure}%
The expected runtime of the \opoea\ on such a problem is in~\bigThetaOf{\scale^{\plateauWidthRaw}}~\cite{AD2018PRAFP}.
The \plateau\ problems are no bijective transformation of \onemax.
Instead, they reduce the number of possible objective values \mbox{($|\mathbb{Z}|<|\mathbb{Y}|$)}.
We can expect that the fitness of the solutions on the plateau will get worse quickly under FFA.
We conduct the same experiment as for the \jump\ function with the \plateau\ function and plot the results in the same manner in Figure~\ref{fig:plateau_runtime}.
This time, the \opofeagz\ performs worse than the \opoea.
Interestingly, if we divide the observed mean runtimes of \opofeagz\ by the problem dimension~\scale, we approximately obtain those observed with \opoeagz\ (see the gray marks in Figure~\ref{fig:plateau_runtime}).
This might be a coincidence and more research is necessary.%
\subsection{Bijection Invariance: Md5 Checksum of Objective Values}%
\label{sec:md5}%
We now repeat our experiments with the \opofeagz\ on the \onemax, \twomax, \leadingones, and \trap\ problems \mbox{with~$\scale\in\intFromTo{3}{32}$}, but use a transformation of the objective functions:
Instead of working on the objective values directly, we optimize their \mbox{Md5~checksums}.
We therefore implement~\ffaH\ as a hash table where their encounter frequencies are stored.
The \mbox{Md5~checksum} is a 128~bit message digest published in~\cite{R1992TMMDA}, where \emph{it is conjectured that it is computationally infeasible to produce two messages having the same message digest}.
Although \mbox{Md5~checksums} are not an encryption method, they do allow us to further test the invariance under such ``extreme'' transformations and the idea of implementing~\ffaH\ as hash table without further assumptions.\footnote{%
It can be assumed that applying \opoea\ to this problem would yield the worst-case complexity and we thus omit doing it.}

We use the same random seeds as in the original runs working on the objective values.
We find that all 3333~runs on all the instances have the same \mbox{(FE, objective value)}-traces as their counterparts, which also follows from Theorem~\ref{theo:invariance}.
Illustrating these results here has no merit, as the figures would be identical to those already shown.
We include the full log files as well as the algorithm implementation in our dataset~\cite{WWLC2019TEDFTSFFAMOAIUBTOTOF}.%
\subsection{\wmodel~Instances}%
The \wmodel~\cite{WW2018DFOCOPATTWMBPFST,W2018TWMATBBDOBPIFTBGW,WNSRG2008ATMFMOERANFL} is a benchmark problem which exhibits different difficult fitness landscape features in a tunable fashion.%
\footnote{%
There was a mistake in~\cite{WW2018DFOCOPATTWMBPFST}: at line~19 of Algorithm~1, ``start'' should be replaced with ``\wmn''. This was corrected in~\cite{WCLW2019SADSOBIFATMPFBBDOA,W2018RFSSAOTWMFBBOB} and was always correct in the \mbox{\wmodel~implementation~\cite{W2018TWMATBBDOBPIFTBGW}}.%
}
These include the base size (via parameter~\wmn), neutrality (via parameter~\wmm), epistasis (via parameter~\wmnu), and ruggedness (via parameter~\wmg), from which instances of dimension~\mbox{$\scale=\wmmn$} result.
The \wmodel\ base problem is equivalent to \onemax\ but searches for a string of alternating 0 and 1~bits.
Different transformations are applied to it.
While the ruggedness transformation is a bijective transformation of objective function, the mappings introducing neutrality and epistasis transform the search space itself.
19~diverse \wmodel\ instances have been selected in~\cite{WCLW2019SADSOBIFATMPFBBDOA} based on a large-scale experiment.
No theoretical bounds for the runtimes on these instances are known, but they exhibit different degrees of empirical hardness for different algorithms.%
\begin{table}[tb]%
\centering%
\caption{%
The \ert\ and fraction~\successFrac\ of the~71 \opoeagz\ runs discovering the optimum on the 19~\wmodel\ problem instances selected in~\cite{WCLW2019SADSOBIFATMPFBBDOA}. As all runs of \opofeagz\ reached the optimum, we present its mean and median runtime.}%
\label{tbl:wmodelresults}%
\clipbox{0pt}{%
\resizebox{\linewidth}{!}{%
\begin{tabular}{@{~}r@{~~}r@{~~}r@{~~}r@{~~}r|rr|rr@{~}}%
\hline%
\multicolumn{5}{c|}{\wmodel~Instance}&\multicolumn{2}{c|}{\opoeagz}&\multicolumn{2}{c}{\opofeagz}\\%
id&\wmn&\wmm&\wmnu&\wmg&\successFrac&\multicolumn{1}{c|}{\ert}&\multicolumn{1}{c}{\meanRuntime}&\multicolumn{1}{c}{\medianRuntime}\\%
\hline%
1&10&2&6&10&1&5'928&1'090&734\\%
2&10&2&6&18&1&6'605&904&815\\%
3&16&1&5&72&1&9'400&3'646&3'191\\%
4&16&3&9&72&1&864'850&5'856&5'163\\%
5&25&1&23&90&0.66&$5.11\!\cdot\!10^{9}$&3'049&2'218\\%
6&32&1&2&397&0&$+\infty$&1'602&1'355\\%
7&32&4&11&0&0.31&$2.23\!\cdot\!10^{10}$&279'944&238'904\\%
8&32&4&14&0&0.31&$2.23\!\cdot\!10^{10}$&287'286&231'266\\%
9&32&4&8&128&0.75&$3.61\!\cdot\!10^{9}$&219'939&201'524\\%
10&50&1&36&245&0.35&$1.84\!\cdot\!10^{10}$&68'248&60'347\\%
11&50&2&21&256&0.46&$1.23\!\cdot\!10^{10}$&572'874&484'795\\%
12&50&3&16&613&0.24&$3.18\!\cdot\!10^{10}$&639'914&568'120\\%
13&64&2&32&256&0.28&$2.55\!\cdot\!10^{10}$&383'998&359'452\\%
14&64&3&21&16&0.27&$2.74\!\cdot\!10^{10}$&$1.00\!\cdot\!10^{6}$&851'246\\%
15&64&3&21&256&0.17&$4.92\!\cdot\!10^{10}$&$1.28\!\cdot\!10^{6}$&$1.07\!\cdot\!10^{6}$\\%
16&64&3&21&403&0.23&$3.44\!\cdot\!10^{10}$&$1.12\!\cdot\!10^{6}$&884'679\\%
17&64&4&52&2&0.42&$1.37\!\cdot\!10^{10}$&612'610&537'448\\%
18&75&1&60&16&0.27&$2.74\!\cdot\!10^{10}$&225'489&184'933\\%
19&75&2&32&4&0.25&$2.94\!\cdot\!10^{10}$&$1.83\!\cdot\!10^{6}$&$1.61\!\cdot\!10^{6}$\\%
\hline%
\end{tabular}%
}}%
\end{table}

We conduct 71~runs for both algorithms on each of these 19~\wmodel\ instances.
In Table~\ref{tbl:wmodelresults}, we presented the fraction~\successFrac\ of runs that found the global optimum and the \ert\ for \opoeagz.
While it can always solve the four easiest instances, its success rate within the $10^{10}$~FEs then drops, which leads to very high \ert~values.
The \opofeagz\ is always faster than \opoeagz\ and all of its runs discovered the global optima of their respective \wmodel\ instances.
In this case, $\meanRuntime=\ert$ and we list it alongside the median runtime~\medianRuntime, which, like on the previously investigated problems, is always smaller than the mean.

Of special interest here is instance~6, which could not be solved by \opoeagz\ at all.
Here, $\scale=\wmmn=32$ and only a ruggedness transformation with $\wmg=397$ is performed, while no additional epistasis ($\wmnu=2$) or neutrality ($\wmm=1$) are introduced in the landscape.
In other words, here, the objective function is equivalent to a (bijective) permutation of the objective values produced by a \onemax\ instance (with a different but equivalent base problem).

This permutation leads to a long deceptive slope in the mid-range of the original objective values and three extremely rugged spikes near the global optimum, i.e., we can expect it to have a hardness similar to the \jump\ or \trap\ functions for the \opoea, which the experiment confirms.
Only for this instance, we conduct 3333 runs with \opofeagz\ and find that the mean~1602 and median~1355 of the runtime are very close to those on the \onemax\ (1620, 1375) and \trap\ functions (1620, 1390), which again confirms the invariance of FFA towards bijective transformations of the objective function.%
\subsection{\maxsat~Problems}%
\label{sec:maxsat}%
The Satisfiability Problem is one of the most prominent problems in artificial intelligence.
An instance is a formula~\mbox{$\maxSatFormula:\booleans^{\scale}\mapsto\booleans$} over~\scale\ Boolean variables.
The variables appear as literals either directly or negated in~\maxSatClauses\ ``{\texttt{or}}'' clauses, which are all combined into one ``{\texttt{and}}''.
Solving a Satisfiability Problem means finding a setting~\solspel\ for the variables so that \maxSatFormulab{\solspel} becomes \texttt{true} (or whether such a setting exists).
This \npHard~\cite{G1979CAIAGTTTONC} decision problem is transformed to an optimization version, the \maxsat\ problem~\cite{HS2005SLSFAA}, where the objective function \ofelb{\solspel}, subject to minimization, computes the number of clauses which are \texttt{false} under~\solspel.
If~\mbox{$\ofelb{\solspel}=0$}, all clauses are \texttt{true}, which solves the Satisfiability Problem.
The worst possible value~\upperBound\ that~\ofel\ can take on is~\maxSatClauses.

The \maxsat\ problem exhibits low epistasis but deceptiveness~\cite{RW1998GABITMD}.
In the so-called phase transition region with~\mbox{$\maxSatClauses/\maxSatVariables\approx4.26$}, the average instance hardness for stochastic local search algorithms is maximal~\cite{HS2000SAORFROS,DNS2017TCAOEAORSkCF,DNS2015IRBFTOPOEOR3CFBOFDC}.
We apply our algorithms as incomplete solvers~\cite{GKSS2008SS} on the ten sets of \emph{satisfiable} uniform random \mbox{3-SAT} instances from SATLib~\cite{HS2000SAORFROS}, which stem from this region.
Here, the number of variables~\scale\ is \mbox{from~$\left\{20\right\}\cup\left\{25i:i\in \intFromTo{2}{10}\right\}$}, where 1000~instances are given for \mbox{$\scale\in\left\{20,50,100\right\}$} and 100 otherwise.
With the \opoeagz, we can only conduct 11~runs for each \mbox{$\scale\in\left\{20,50,75\right\}$} due to the high runtime requirement resulting from many runs failing to solve the problem within \mbox{$10^{10}$~FEs}.
With the \opofeagz, we conduct 11~runs for \mbox{$\scale\in\left\{20,50,100\right\}$} and 110~runs for each dimension other than these, i.e., have \mbox{$110*100=11*1000=11'000$~runs} for each instance dimension~\scale\ in SATLib.%
\begin{table}[tb]%
\caption{The fraction~\successFrac\ of successful runs, the \ert, and the mean end objective value~\meanBestF\ for \opoeagz\ and \opoeagz\ on the satisfiable \maxsat\ instances from SATLib.}%
\label{tbl:maxsat}%
\centering%
\resizebox{\linewidth}{!}{%
\begin{tabular}{r|r@{~~}r@{~}r|r@{~~}r@{~}r}%
\hline%
\multicolumn{1}{c|}{instance}&\multicolumn{3}{c|}{\opoeagz}&\multicolumn{3}{c}{\opofeagz}\\%
\multicolumn{1}{c|}{set}&\multicolumn{1}{c}{\successFrac}&\multicolumn{1}{c}{\ert}&\multicolumn{1}{c|}{\meanBestF}&\multicolumn{1}{c}{\successFrac}&\multicolumn{1}{c}{\ert}&\multicolumn{1}{c}{\meanBestF}\\%
\hline%
\instance{uf20\_*}&0.985&$1.91\!\cdot\!10^{8}$&0.0154&1&3'091&0\\%
\instance{uf50\_*}&0.748&$3.56\!\cdot\!10^{9}$&0.299&1&93'459&0\\%
\instance{uf75\_*}&0.583&$7.41\!\cdot\!10^{9}$&0.528&1&490'166&0\\%
\instance{uf100\_*}&\multicolumn{1}{c}{--}&\multicolumn{1}{c}{--}&\multicolumn{1}{c|}{--}&1&$2.14\!\cdot\!10^{6}$&0\\%
\instance{uf125\_*}&\multicolumn{1}{c}{--}&\multicolumn{1}{c}{--}&\multicolumn{1}{c|}{--}&1&$5.27\!\cdot\!10^{6}$&0\\%
\instance{uf150\_*}&\multicolumn{1}{c}{--}&\multicolumn{1}{c}{--}&\multicolumn{1}{c|}{--}&1&$1.40\!\cdot\!10^{7}$&0\\%
\instance{uf175\_*}&\multicolumn{1}{c}{--}&\multicolumn{1}{c}{--}&\multicolumn{1}{c|}{--}&1&$5.78\!\cdot\!10^{7}$&0\\%
\instance{uf200\_*}&\multicolumn{1}{c}{--}&\multicolumn{1}{c}{--}&\multicolumn{1}{c|}{--}&0.991&$2.44\!\cdot\!10^{8}$&0.00945\\%
\instance{uf225\_*}&\multicolumn{1}{c}{--}&\multicolumn{1}{c}{--}&\multicolumn{1}{c|}{--}&0.994&$2.43\!\cdot\!10^{8}$&0.00555\\%
\instance{uf250\_*}&\multicolumn{1}{c}{--}&\multicolumn{1}{c}{--}&\multicolumn{1}{c|}{--}&0.992&$2.43\!\cdot\!10^{8}$&0.00782\\%
\hline%
\end{tabular}%
}%
\end{table}

The overall performance of the algorithms aggregated over the instance sets is given in Table~\ref{tbl:maxsat}.
We find that the \opofeagz\ performs much better than the \opoeagz.
While the former can reliably solve instances of all dimensions, the latter already fails in almost half of the runs \mbox{for~$\scale=75$}.
The overall \ert\ of the \opofeagz\ for \mbox{dimension~$\scale=250$} is only about 7\% of the \ert\ that the \opoeagz\ needs over all instances of \mbox{$\scale=50$}.

\begin{figure}[tb]%
\centering%
\includegraphics[width=0.9999\linewidth]{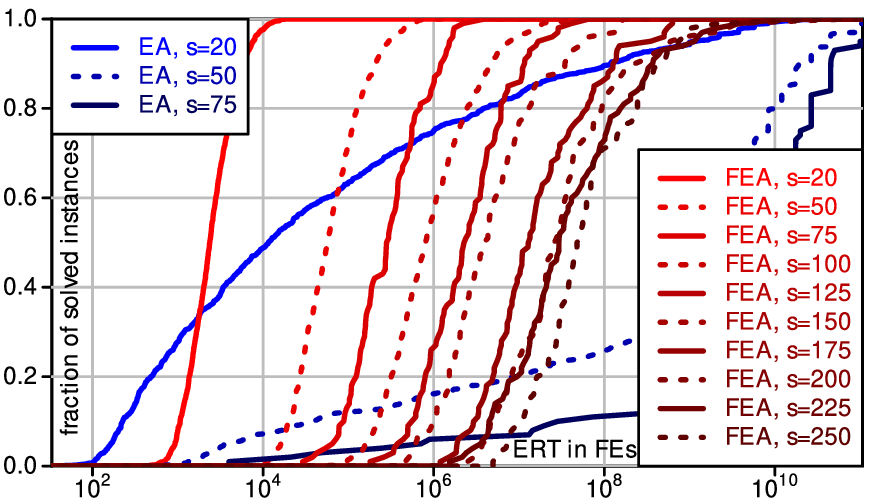}%
\caption{The ERT-ECDF curves for the SATLib instances: the fraction of instances of a given dimension~\scale\ solved over their empirically determined expected runtime.}%
\label{fig:maxsat_ert_ecdf}%
\end{figure}

We now plot the Empirical Cumulative Distribution Function (ECDF~\cite{HAFR2012RPBBOBES}) over the estimated \ert\ in Figure~\ref{fig:maxsat_ert_ecdf}.
Normally, the ECDF shows the fraction of \emph{runs} that could solve their corresponding problem instance over time.
However, we want to illustrate which algorithm can solve which fraction of the \emph{instances} until which (empirically determined expected) time.
For a given dimension~\scale, we therefore compute the \ert\ for each of the corresponding instances based on the conducted runs.

It seems that SATLib contains some instances that the \opoeagz\ can solve quickly, but on many instances it is slow or fails often.
The \ert\ of the instance of dimension~$\scale=250$ hardest for the \opofeagz\ is only~38\% higher than the \ert\ of the \mbox{scale-20} instance hardest for the \opoeagz.
Due to the drop in success rate, the behavior of the \opoeagz\ is already very unstable at dimensions~50 and~75.
This does not happen for the \opofeagz\ at any of the tested dimensions.

In summary, the \opofea\ very significantly outperforms the \opoea\ on a practically-relevant task, which goes beyond the scope of toy problems.%
\subsection{Job Shop Scheduling Problems (\jssp)}%
\label{sec:jssp}%
With the \maxsat, we have investigated an important \npHard\ problem.
While exhibiting interesting features, the \opofeagz\ algorithm we applied is not competitive to the state-of-the-art even two decades ago~\cite{HS2000LSAFSAEE}.
We now want to investigate if FFA can be helpful when the base algorithm is already performing well and we will do so on an entirely different domain.

In a job shop scheduling problem (\jssp)~\cite{LLRKS1993SASAAC} without preemption, there are \jsspMachines~machines and \jsspJobs~jobs.
Each job must be processed by all machines in a job-specific sequence and has, for each machine, a specific processing time.
The goal is to find assignments of jobs to machines that result in an overall shortest makespan, i.e., the schedule which can complete all the jobs the fastest.
The \jssp\ is \npHard~\cite{LLRKS1993SASAAC}.
The objective values are positive integers since the processing times are integers.
We obtain an upper bound~\upperBound\ needed for FFA as the sum of all processing times of all sub-jobs.
We use the \jssp\ as educational example in~\cite{W2019AITOA}, where we discuss all of the following components (except FFA) in great detail.

A solution for the \jssp\ is encoded as permutation with repetition, as integer strings where each of the \jsspJobs~job IDs occurs exactly \jsspMachines~times~\cite{GTK1994SJSSPBGA}.
Such an integer string~\solspel\ is processed from front to end.
When encountering job~$i$, we know to which machine~$j$ it needs to go next based on the job-specific machine sequence and on how often we already saw~$i$ in~\solspel\ before.
We can start it on~$j$ at a time which is the maximum of \emph{1)}~when the previous sub-job assigned to~$j$ will finish and \emph{2)}~when the previous sub-job of~$i$ completes on its corresponding machine.

We develop a memetic algorithm~\cite{NCM2012HOMA} which retains the~\mbox{$\paramMu=16$} best candidate solutions in its population and generates \mbox{$\paramLambda=16$}~new strings in each step via recombination.
Recombination proceeds similar to the solution decoding, but reads unprocessed sub-jobs iteratively from two parent strings (between which it randomly switches) and writes them to an offspring, while marking each processed sub-job in both parents as processed~\cite{W2019AITOA}.
The \paramLambda~new strings each are refined with ten steps of a local search which, in each step, scans the single-swap neighborhood of the string in random order until it finds a makespan-improving move and applies it (or stops if none can be found).

The two algorithms we investigate differ \emph{only} in what they do once this step is completed:
The first, \jsspMA, now applies selection based on the objective values.
In the \jsspFFAMA, on the other hand, the FFA table \ffaH\ is updated by increasing the frequency counter of the corresponding objective value of each of the \mbox{$\paramMu+\paramLambda$} solutions in the joint parent-offspring population.
Selection chooses the $\paramMu$~solutions with the lowest frequency fitness value.
\jsspFFAMA\ still uses the objective function~\ofel\ in the local search and to break ties in FFA. It is therefore not invariant under bijections of~\ofel.

Our goal this time is to achieve the best possible result within five minutes of runtime on an Intel Core \mbox{i7~8700} CPU with \mbox{3.2~GHz} and \mbox{16~GiB} RAM under Java OpenJDK~13 on~\mbox{Ubuntu~19.04}.
This is very different from the previous goals of solving the problems to optimality.
We conduct~$\runs=11$ runs each on 82~well-known \jssp\ instances from the \mbox{OR-Library}~\cite{B1990OLDTPBEM,B1990OLJSS}, namely the sets %
\instance{abz*}, %
\instance{ft*}, %
\instance{la*}, %
\instance{orb*}, %
\instance{swv*}, and %
\instance{yn*}, where all processing times are integers.%
\begin{table}%
\caption{%
Best, median, mean, and standard deviation of results from 11~runs of \jsspMA\ and \jsspFFAMA; \tblbettertxt: better value, \tblsolvedtxt: best known solutions (BKS) reached.%
}%
\label{tbl:jssp:results}%
\centering%
\clipbox{0pt}{%
\resizebox{\linewidth}{!}{%
\begin{tabular}{@{\hspace{0.17em}}l@{~}r@{~}l|rrrr|rrrr@{\hspace{0.17em}}}%
\hline%
&&&\multicolumn{4}{c|}{\jsspMA}&\multicolumn{4}{c}{\jsspFFAMA}\\%
\multicolumn{1}{l}{inst}&\multicolumn{2}{l|}{BKS+ref}&\multicolumn{1}{c}{best}&\multicolumn{1}{c}{med}&\multicolumn{1}{c}{mean}&\multicolumn{1}{c|}{sd}&\multicolumn{1}{c}{best}&\multicolumn{1}{c}{med}&\multicolumn{1}{c}{mean}&\multicolumn{1}{c}{sd}\\%
\hline%
\instance{abz5}&1234&\cite{AC1991ACSOTJSSP}&1239&1244&1244.1&4.9&\tblsolved{\tblbetter{1234}}&\tblsolved{\tblbetter{1234}}&\tblsolved{\tblbetter{1236.2}}&\tblsolved{2.5}\\%
\instance{abz6}&943&\cite{AC1991ACSOTJSSP}&947&948&949.0&3.9&\tblsolved{\tblbetter{943}}&\tblsolved{\tblbetter{943}}&\tblsolved{\tblbetter{943.4}}&\tblsolved{1.2}\\%
\instance{abz7}&656&\cite{H2002PJSSP}&\tblbetter{679}&693&694.7&10.8&685&\tblbetter{691}&\tblbetter{693.1}&5.5\\%
\instance{abz8}&665&\cite{H2002PJSSP}&698&709&713.2&12.4&\tblbetter{688}&\tblbetter{706}&\tblbetter{705.9}&8.2\\%
\instance{abz9}&678&\cite{ZSRQ2008SNROTSAATTJSSP}&724&737&736.2&6.7&\tblbetter{714}&\tblbetter{727}&\tblbetter{725.6}&6.7\\%
\hline%
\instance{ft06}&55&\cite{CP1989AAFSTJSP}&\tblsolvedReliably{55}&\tblsolvedReliably{55}&\tblsolvedReliably{55.0}&\tblsolvedReliably{0.0}&\tblsolvedReliably{55}&\tblsolvedReliably{55}&\tblsolvedReliably{55.0}&\tblsolvedReliably{0.0}\\%
\instance{ft10}&930&\cite{CP1989AAFSTJSP}&937&949&948.2&7.6&\tblsolved{\tblbetter{930}}&\tblsolved{\tblbetter{930}}&\tblsolved{\tblbetter{933.9}}&\tblsolved{6.1}\\%
\instance{ft20}&1165&\cite{CP1989AAFSTJSP}&1173&1178&1177.9&1.8&\tblsolved{\tblbetter{1165}}&\tblsolved{1178}&\tblsolved{\tblbetter{1176.4}}&\tblsolved{4.1}\\%
\hline%
\instance{la01}&666&\cite{AC1991ACSOTJSSP}&\tblsolvedReliably{666}&\tblsolvedReliably{666}&\tblsolvedReliably{666.0}&\tblsolvedReliably{0.0}&\tblsolvedReliably{666}&\tblsolvedReliably{666}&\tblsolvedReliably{666.0}&\tblsolvedReliably{0.0}\\%
\instance{la02}&655&\cite{AC1991ACSOTJSSP}&\tblsolvedReliably{655}&\tblsolvedReliably{655}&\tblsolvedReliably{655.0}&\tblsolvedReliably{0.0}&\tblsolvedReliably{655}&\tblsolvedReliably{655}&\tblsolvedReliably{655.0}&\tblsolvedReliably{0.0}\\%
\instance{la03}&597&\cite{AC1991ACSOTJSSP}&\tblsolved{597}&\tblsolved{597}&\tblsolved{599.2}&\tblsolved{4.9}&\tblsolvedReliably{597}&\tblsolvedReliably{597}&\tblsolvedReliably{\tblbetter{597.0}}&\tblsolvedReliably{0.0}\\%
\instance{la04}&590&\cite{AC1991ACSOTJSSP}&\tblsolvedReliably{590}&\tblsolvedReliably{590}&\tblsolvedReliably{590.0}&\tblsolvedReliably{0.0}&\tblsolvedReliably{590}&\tblsolvedReliably{590}&\tblsolvedReliably{590.0}&\tblsolvedReliably{0.0}\\%
\instance{la05}&593&\cite{AC1991ACSOTJSSP}&\tblsolvedReliably{593}&\tblsolvedReliably{593}&\tblsolvedReliably{593.0}&\tblsolvedReliably{0.0}&\tblsolvedReliably{593}&\tblsolvedReliably{593}&\tblsolvedReliably{593.0}&\tblsolvedReliably{0.0}\\%
\instance{la06}&926&\cite{AC1991ACSOTJSSP}&\tblsolvedReliably{926}&\tblsolvedReliably{926}&\tblsolvedReliably{926.0}&\tblsolvedReliably{0.0}&\tblsolvedReliably{926}&\tblsolvedReliably{926}&\tblsolvedReliably{926.0}&\tblsolvedReliably{0.0}\\%
\instance{la07}&890&\cite{AC1991ACSOTJSSP}&\tblsolvedReliably{890}&\tblsolvedReliably{890}&\tblsolvedReliably{890.0}&\tblsolvedReliably{0.0}&\tblsolvedReliably{890}&\tblsolvedReliably{890}&\tblsolvedReliably{890.0}&\tblsolvedReliably{0.0}\\%
\instance{la08}&863&\cite{AC1991ACSOTJSSP}&\tblsolvedReliably{863}&\tblsolvedReliably{863}&\tblsolvedReliably{863.0}&\tblsolvedReliably{0.0}&\tblsolvedReliably{863}&\tblsolvedReliably{863}&\tblsolvedReliably{863.0}&\tblsolvedReliably{0.0}\\%
\instance{la09}&951&\cite{AC1991ACSOTJSSP}&\tblsolvedReliably{951}&\tblsolvedReliably{951}&\tblsolvedReliably{951.0}&\tblsolvedReliably{0.0}&\tblsolvedReliably{951}&\tblsolvedReliably{951}&\tblsolvedReliably{951.0}&\tblsolvedReliably{0.0}\\%
\instance{la10}&958&\cite{AC1991ACSOTJSSP}&\tblsolvedReliably{958}&\tblsolvedReliably{958}&\tblsolvedReliably{958.0}&\tblsolvedReliably{0.0}&\tblsolvedReliably{958}&\tblsolvedReliably{958}&\tblsolvedReliably{958.0}&\tblsolvedReliably{0.0}\\%
\hline%
\instance{la11}&1222&\cite{AC1991ACSOTJSSP}&\tblsolvedReliably{1222}&\tblsolvedReliably{1222}&\tblsolvedReliably{1222.0}&\tblsolvedReliably{0.0}&\tblsolvedReliably{1222}&\tblsolvedReliably{1222}&\tblsolvedReliably{1222.0}&\tblsolvedReliably{0.0}\\%
\instance{la12}&1039&\cite{AC1991ACSOTJSSP}&\tblsolvedReliably{1039}&\tblsolvedReliably{1039}&\tblsolvedReliably{1039.0}&\tblsolvedReliably{0.0}&\tblsolvedReliably{1039}&\tblsolvedReliably{1039}&\tblsolvedReliably{1039.0}&\tblsolvedReliably{0.0}\\%
\instance{la13}&1150&\cite{AC1991ACSOTJSSP}&\tblsolvedReliably{1150}&\tblsolvedReliably{1150}&\tblsolvedReliably{1150.0}&\tblsolvedReliably{0.0}&\tblsolvedReliably{1150}&\tblsolvedReliably{1150}&\tblsolvedReliably{1150.0}&\tblsolvedReliably{0.0}\\%
\instance{la14}&1292&\cite{AC1991ACSOTJSSP}&\tblsolvedReliably{1292}&\tblsolvedReliably{1292}&\tblsolvedReliably{1292.0}&\tblsolvedReliably{0.0}&\tblsolvedReliably{1292}&\tblsolvedReliably{1292}&\tblsolvedReliably{1292.0}&\tblsolvedReliably{0.0}\\%
\instance{la15}&1207&\cite{AC1991ACSOTJSSP}&\tblsolvedReliably{1207}&\tblsolvedReliably{1207}&\tblsolvedReliably{1207.0}&\tblsolvedReliably{0.0}&\tblsolvedReliably{1207}&\tblsolvedReliably{1207}&\tblsolvedReliably{1207.0}&\tblsolvedReliably{0.0}\\%
\instance{la16}&945&\cite{AC1991ACSOTJSSP}&946&946&959.4&17.5&\tblsolved{\tblbetter{945}}&\tblsolved{946}&\tblsolved{\tblbetter{945.9}}&\tblsolved{0.3}\\%
\instance{la17}&784&\cite{AC1991ACSOTJSSP}&\tblsolved{784}&\tblsolved{787}&\tblsolved{787.5}&\tblsolved{3.0}&\tblsolvedReliably{784}&\tblsolvedReliably{\tblbetter{784}}&\tblsolvedReliably{\tblbetter{784.0}}&\tblsolvedReliably{0.0}\\%
\instance{la18}&848&\cite{AC1991ACSOTJSSP}&\tblsolved{848}&\tblsolved{848}&\tblsolved{850.0}&\tblsolved{4.5}&\tblsolvedReliably{848}&\tblsolvedReliably{848}&\tblsolvedReliably{\tblbetter{848.0}}&\tblsolvedReliably{0.0}\\%
\instance{la19}&842&\cite{AC1991ACSOTJSSP}&\tblsolved{842}&\tblsolved{852}&\tblsolved{853.4}&\tblsolved{8.7}&\tblsolved{842}&\tblsolved{\tblbetter{842}}&\tblsolved{\tblbetter{842.9}}&\tblsolved{3.0}\\%
\instance{la20}&902&\cite{AC1991ACSOTJSSP}&907&907&907.8&1.8&\tblsolved{\tblbetter{902}}&\tblsolved{907}&\tblsolved{\tblbetter{906.5}}&\tblsolved{1.5}\\%
\hline%
\instance{la21}&1046&\cite{YN1997GAFJSSP}&1056&1068&1067.8&7.5&\tblbetter{1047}&\tblbetter{1053}&\tblbetter{1052.5}&2.7\\%
\instance{la22}&927&\cite{AC1991ACSOTJSSP}&935&941&941.3&8.0&\tblbetter{930}&\tblbetter{935}&\tblbetter{934.5}&1.9\\%
\instance{la23}&1032&\cite{AC1991ACSOTJSSP}&\tblsolvedReliably{1032}&\tblsolvedReliably{1032}&\tblsolvedReliably{1032.0}&\tblsolvedReliably{0.0}&\tblsolvedReliably{1032}&\tblsolvedReliably{1032}&\tblsolvedReliably{1032.0}&\tblsolvedReliably{0.0}\\%
\instance{la24}&935&\cite{AC1991ACSOTJSSP}&941&964&960.2&11.7&941&\tblbetter{946}&\tblbetter{945.6}&3.2\\%
\instance{la25}&977&\cite{AC1991ACSOTJSSP}&986&998&1002.3&14.8&\tblbetter{984}&\tblbetter{986}&\tblbetter{986.7}&3.1\\%
\instance{la26}&1218&\cite{AC1991ACSOTJSSP}&\tblsolved{1218}&\tblsolved{1218}&\tblsolved{1222.4}&\tblsolved{11.6}&\tblsolvedReliably{1218}&\tblsolvedReliably{1218}&\tblsolvedReliably{\tblbetter{1218.0}}&\tblsolvedReliably{0.0}\\%
\instance{la27}&1235&\cite{YN1997GAFJSSP}&1252&1269&1268.3&8.3&\tblbetter{1248}&\tblbetter{1264}&\tblbetter{1264.0}&6.7\\%
\instance{la28}&1216&\cite{AC1991ACSOTJSSP}&1225&1232&1238.7&14.5&\tblsolved{\tblbetter{1216}}&\tblsolved{\tblbetter{1225}}&\tblsolved{\tblbetter{1228.1}}&\tblsolved{8.8}\\%
\instance{la29}&1152&\cite{H2002PJSSP}&1199&1222&1224.2&18.9&\tblbetter{1191}&\tblbetter{1219}&\tblbetter{1212.6}&13.0\\%
\instance{la30}&1355&\cite{AC1991ACSOTJSSP}&\tblsolvedReliably{1355}&\tblsolvedReliably{1355}&\tblsolvedReliably{1355.0}&\tblsolvedReliably{0.0}&\tblsolvedReliably{1355}&\tblsolvedReliably{1355}&\tblsolvedReliably{1355.0}&\tblsolvedReliably{0.0}\\%
\hline%
\instance{la31}&1784&\cite{AC1991ACSOTJSSP}&\tblsolvedReliably{1784}&\tblsolvedReliably{1784}&\tblsolvedReliably{1784.0}&\tblsolvedReliably{0.0}&\tblsolvedReliably{1784}&\tblsolvedReliably{1784}&\tblsolvedReliably{1784.0}&\tblsolvedReliably{0.0}\\%
\instance{la32}&1850&\cite{AC1991ACSOTJSSP}&\tblsolvedReliably{1850}&\tblsolvedReliably{1850}&\tblsolvedReliably{1850.0}&\tblsolvedReliably{0.0}&\tblsolvedReliably{1850}&\tblsolvedReliably{1850}&\tblsolvedReliably{1850.0}&\tblsolvedReliably{0.0}\\%
\instance{la33}&1719&\cite{AC1991ACSOTJSSP}&\tblsolvedReliably{1719}&\tblsolvedReliably{1719}&\tblsolvedReliably{1719.0}&\tblsolvedReliably{0.0}&\tblsolvedReliably{1719}&\tblsolvedReliably{1719}&\tblsolvedReliably{1719.0}&\tblsolvedReliably{0.0}\\%
\instance{la34}&1721&\cite{AC1991ACSOTJSSP}&\tblsolvedReliably{1721}&\tblsolvedReliably{1721}&\tblsolvedReliably{1721.0}&\tblsolvedReliably{0.0}&\tblsolvedReliably{1721}&\tblsolvedReliably{1721}&\tblsolvedReliably{1721.0}&\tblsolvedReliably{0.0}\\%
\instance{la35}&1888&\cite{AC1991ACSOTJSSP}&\tblsolvedReliably{1888}&\tblsolvedReliably{1888}&\tblsolvedReliably{1888.0}&\tblsolvedReliably{0.0}&\tblsolvedReliably{1888}&\tblsolvedReliably{1888}&\tblsolvedReliably{1888.0}&\tblsolvedReliably{0.0}\\%
\instance{la36}&1268&\cite{AC1991ACSOTJSSP}&1295&1301&1307.7&15.7&\tblbetter{1281}&\tblbetter{1297}&\tblbetter{1297.5}&9.8\\%
\instance{la37}&1397&\cite{AC1991ACSOTJSSP}&1446&1467&1462.3&13.5&\tblbetter{1432}&\tblbetter{1446}&\tblbetter{1442.5}&6.3\\%
\instance{la38}&1196&\cite{NS1996AFTSAFTJSP}&1251&1263&1262.9&13.5&\tblbetter{1239}&\tblbetter{1240}&\tblbetter{1244.6}&8.7\\%
\instance{la39}&1233&\cite{AC1991ACSOTJSSP}&1251&1256&1267.0&20.5&\tblbetter{1248}&\tblbetter{1250}&\tblbetter{1250.0}&1.2\\%
\instance{la40}&1222&\cite{AC1991ACSOTJSSP}&1241&1264&1262.1&14.0&\tblbetter{1233}&\tblbetter{1247}&\tblbetter{1247.3}&7.7\\%
\hline%
\instance{orb01}&1059&\cite{AC1991ACSOTJSSP}&\tblsolved{\tblbetter{1059}}&\tblsolved{1104}&\tblsolved{1099.2}&\tblsolved{17.5}&1071&\tblbetter{1071}&\tblbetter{1075.4}&5.6\\%
\instance{orb02}&888&\cite{AC1991ACSOTJSSP}&890&919&909.0&14.3&\tblbetter{889}&\tblbetter{889}&\tblbetter{889.0}&0.0\\%
\instance{orb03}&1005&\cite{AC1991ACSOTJSSP}&1026&1058&1060.5&27.8&\tblsolved{\tblbetter{1005}}&\tblsolved{\tblbetter{1022}}&\tblsolved{\tblbetter{1019.5}}&\tblsolved{7.4}\\%
\instance{orb04}&1005&\cite{AC1991ACSOTJSSP}&\tblsolved{1005}&\tblsolved{1028}&\tblsolved{1024.3}&\tblsolved{14.0}&\tblsolved{1005}&\tblsolved{\tblbetter{1011}}&\tblsolved{\tblbetter{1010.0}}&\tblsolved{2.2}\\%
\instance{orb05}&887&\cite{AC1991ACSOTJSSP}&890&905&913.8&18.1&\tblsolved{\tblbetter{887}}&\tblsolved{\tblbetter{890}}&\tblsolved{\tblbetter{890.2}}&\tblsolved{2.4}\\%
\instance{orb06}&1010&\cite{BV1998GLSWSBFJSS}&1013&1031&1031.0&8.6&1013&\tblbetter{1013}&\tblbetter{1017.8}&6.0\\%
\instance{orb07}&397&\cite{H2002PJSSP}&\tblsolved{397}&\tblsolved{397}&\tblsolved{401.7}&\tblsolved{7.4}&\tblsolved{397}&\tblsolved{397}&\tblsolved{\tblbetter{397.1}}&\tblsolved{0.3}\\%
\instance{orb08}&899&\cite{BV1998GLSWSBFJSS}&914&944&941.3&16.5&\tblsolved{\tblbetter{899}}&\tblsolved{\tblbetter{899}}&\tblsolved{\tblbetter{902.0}}&\tblsolved{5.4}\\%
\instance{orb09}&934&\cite{BV1998GLSWSBFJSS}&939&945&947.9&7.2&\tblsolved{\tblbetter{934}}&\tblsolved{\tblbetter{939}}&\tblsolved{\tblbetter{937.9}}&\tblsolved{3.4}\\%
\instance{orb10}&944&\cite{BV1998GLSWSBFJSS}&\tblsolved{944}&\tblsolved{946}&\tblsolved{957.3}&\tblsolved{15.9}&\tblsolvedReliably{944}&\tblsolvedReliably{\tblbetter{944}}&\tblsolvedReliably{\tblbetter{944.0}}&\tblsolvedReliably{0.0}\\%
\hline%
\instance{swv01}&1407&\cite{H2002PJSSP}&1476&1517&1524.1&35.1&\tblbetter{1447}&\tblbetter{1474}&\tblbetter{1483.3}&20.9\\%
\instance{swv02}&1475&\cite{H2002PJSSP}&1550&1585&1582.8&20.9&\tblbetter{1525}&\tblbetter{1548}&\tblbetter{1549.1}&15.5\\%
\instance{swv03}&1398&\cite{H2002PJSSP}&1500&1530&1533.3&28.8&\tblbetter{1489}&\tblbetter{1512}&\tblbetter{1513.1}&15.1\\%
\instance{swv04}&1464&\vphantom{\cite{VLS2015FDSFCBS}}\cite{VLS2015FDSFCBSDER}&1580&1615&1624.1&30.6&\tblbetter{1564}&\tblbetter{1578}&\tblbetter{1586.5}&22.3\\%
\instance{swv05}&1424&\cite{H2002PJSSP}&\tblbetter{1517}&1575&1585.3&49.8&1523&\tblbetter{1554}&\tblbetter{1556.0}&25.8\\%
\instance{swv06}&1671&\cite{VLS2015FDSFCBSDER}&1859&1903&1909.2&44.0&\tblbetter{1824}&\tblbetter{1864}&\tblbetter{1862.7}&27.8\\%
\instance{swv07}&1594&\cite{GR2014AEAGMWABRKGAFJSS}&1766&1814&1816.1&26.4&\tblbetter{1705}&\tblbetter{1755}&\tblbetter{1753.3}&24.5\\%
\instance{swv08}&1752&\cite{VLS2015FDSFCBSDER}&1940&1992&1989.9&38.0&\tblbetter{1930}&\tblbetter{1946}&\tblbetter{1946.4}&13.9\\%
\instance{swv09}&1655&\cite{VLS2015FDSFCBSDER}&1820&1871&1877.9&51.8&\tblbetter{1805}&\tblbetter{1844}&\tblbetter{1844.8}&19.3\\%
\instance{swv10}&1743&\cite{GR2014AEAGMWABRKGAFJSS}&1909&1956&1974.5&44.7&\tblbetter{1904}&\tblbetter{1936}&\tblbetter{1931.4}&15.4\\%
\hline%
\instance{swv11}&2983&\cite{NS2005AATSAFTJSP}&\tblbetter{3439}&\tblbetter{3506}&\tblbetter{3506.1}&52.6&3495&3574&3583.5&62.6\\%
\instance{swv12}&2977&\cite{PLC2015ATSPRATSTJSSP}&\tblbetter{3478}&\tblbetter{3594}&\tblbetter{3594.4}&56.3&3511&3605&3622.1&66.3\\%
\instance{swv13}&3104&\cite{H2002PJSSP}&\tblbetter{3543}&3679&3686.5&81.1&3578&\tblbetter{3664}&\tblbetter{3677.2}&78.6\\%
\instance{swv14}&2968&\cite{H2002PJSSP}&\tblbetter{3358}&3455&\tblbetter{3444.4}&60.4&3369&\tblbetter{3454}&3452.9&69.1\\%
\instance{swv15}&2885&\cite{PLC2015ATSPRATSTJSSP}&3361&\tblbetter{3500}&\tblbetter{3490.6}&89.5&\tblbetter{3356}&3529&3524.1&98.5\\%
\instance{swv16}&2924&\cite{H2002PJSSP}&\tblsolvedReliably{2924}&\tblsolvedReliably{2924}&\tblsolvedReliably{2924.0}&\tblsolvedReliably{0.0}&\tblsolvedReliably{2924}&\tblsolvedReliably{2924}&\tblsolvedReliably{2924.0}&\tblsolvedReliably{0.0}\\%
\instance{swv17}&2794&\cite{H2002PJSSP}&\tblsolvedReliably{2794}&\tblsolvedReliably{2794}&\tblsolvedReliably{2794.0}&\tblsolvedReliably{0.0}&\tblsolvedReliably{2794}&\tblsolvedReliably{2794}&\tblsolvedReliably{2794.0}&\tblsolvedReliably{0.0}\\%
\instance{swv18}&2852&\cite{H2002PJSSP}&\tblsolvedReliably{2852}&\tblsolvedReliably{2852}&\tblsolvedReliably{2852.0}&\tblsolvedReliably{0.0}&\tblsolvedReliably{2852}&\tblsolvedReliably{2852}&\tblsolvedReliably{2852.0}&\tblsolvedReliably{0.0}\\%
\instance{swv19}&2843&\cite{H2002PJSSP}&\tblsolvedReliably{2843}&\tblsolvedReliably{2843}&\tblsolvedReliably{2843.0}&\tblsolvedReliably{0.0}&\tblsolvedReliably{2843}&\tblsolvedReliably{2843}&\tblsolvedReliably{2843.0}&\tblsolvedReliably{0.0}\\%
\instance{swv20}&2823&\cite{H2002PJSSP}&\tblsolvedReliably{2823}&\tblsolvedReliably{2823}&\tblsolvedReliably{2823.0}&\tblsolvedReliably{0.0}&\tblsolvedReliably{2823}&\tblsolvedReliably{2823}&\tblsolvedReliably{2823.0}&\tblsolvedReliably{0.0}\\%
\hline%
\instance{yn1}&884&\cite{ZSRQ2008SNROTSAATTJSSP}&921&934&936.2&11.7&\tblbetter{909}&\tblbetter{931}&\tblbetter{931.5}&10.7\\%
\instance{yn2}&904&\cite{GR2014AEAGMWABRKGAFJSS}&953&962&964.0&7.5&\tblbetter{937}&\tblbetter{954}&\tblbetter{952.4}&9.8\\%
\instance{yn3}&892&\cite{NS2005AATSAFTJSP}&929&951&951.9&16.1&\tblbetter{913}&\tblbetter{938}&\tblbetter{938.6}&12.0\\%
\instance{yn4}&968&\cite{H2002PJSSP}&\tblbetter{1022}&1046&1048.9&20.0&1024&\tblbetter{1041}&\tblbetter{1041.9}&9.0\\%
\hline%
\end{tabular}%
}}%
\end{table}

From Table~\ref{tbl:jssp:results}, we can find that \jsspMA\ can already discover the best known solution (BKS) on 36~instances at least once and always on~27.
\jsspFFAMA, however, can do so 46 and 32~times, respectively.
\jsspFFAMA\ has better best, median, and mean results 37, 45, and 51~times, respectively, while the same is true for the \jsspMA\ only 8, 3, and 4~times.
In other words, on 93\% of the instances that are not already always solved to optimality by \jsspMA, \jsspFFAMA\ has a better mean result.
The mean (median) result of \jsspFFAMA\ is better than the best result of \jsspMA\ in 17 (13)~instances, while the opposite is never true.
\jsspFFAMA\ has a smaller standard deviation in 48~cases, \jsspMA\ only in~4.
We apply the two-sided Welch's \mbox{t-test} to the results on the 49~instances where the algorithms have different mean results with non-zero standard deviations.
\jsspFFAMA\ performs significantly better than \jsspMA\ on 25 of them at a significance level of~$\alpha=0.01$.
Such high significance is a very strong result at only~11 runs.
The opposite is true only on \instance{swv11}, even if we set~$\alpha=0.1$.

Neither \jsspMA\ nor \jsspFFAMA\ can outperform the state-of-the-art on the JSSP, but they are not very far off, at least if we consider result quality only:
The basic \jsspMA\ obtains better best (mean) results than the GWO proposed in~\cite{JZ2018AOGWOFSCPJSAFJSSC} (2018) in 16 (21) of the 39~instances for which results are provided, while the opposite is never (once) true.
While the HFSAQ~\cite{AKZ2016FSAHWQFSJSSP} (2016) has better mean (best) result quality in 23 (12) of 48~comparable cases, \jsspFFAMA\ scores even in the rest, while having better mean solution quality on 4~instances.
It also 13~times achieves better mean makespans (28 times worse ones) on the 63~common instances compared to the HIMGA~\cite{K2015ANHIMGAFJSSP} (2015), while its best solution is never better.
On instances \instance{swv16} to \instance{swv20}, which can be solved to the BKS by both \jsspMA\ and \jsspFFAMA, budgets of more than 16~min were used in~\cite{H2002PJSSP} to find said BKSes.\footnote{%
Of course on older hardware, but our Java implementation is not optimized.%
}
Still, the \jsspFFAMA\ is worse than, e.g., the algorithms in~\cite{CPL2016AHEATSTJSSP}~(2016) and~\cite{VLS2015FDSFCBSDER}~(2015) on every common instance where it does not find the BKS.

In summary, we find that even in a more complicated setup based on an algorithm that already does not perform badly in comparison to recent publications, FFA can lead to a significant performance improvement.
This does not mean that other diversity improvement strategies, e.g., those from Section~\ref{sec:relatedWork}, could not have improved the performance of the \jsspMA\ as well or even better.
Still, together with the results on the \maxsat~problems in Section~\ref{sec:maxsat} and those in our earlier papers on FFA on domains such as Genetic Programming~\cite{WWTY2014EEIAWGP,WWTWDY2014FFA}, this adds evidence to the idea that FFA may not just be of purely academic interest.%
\section{Conclusions}%
\label{sec:conclusions}%
In this paper, we plugged Frequency Fitness Assignment (FFA) into the most basic evolutionary algorithm, the \opoea, and applied the resulting \opofea\ to several problems defined over bit strings of dimension~\scale.
On the one hand, we found that the \opofea\ is slower than the \opoea\ on the \onemax, \leadingones, and \plateau\ functions.
In our experiments with these problems, it seems to increase the mean runtime needed to discover the global optimum by a factor no worse than linear in the number of objective values or in~\scale.
On the other hand, FFA can seemingly decrease the mean runtime on the \trap, \jump, and \twomax\ problems from exponential to the scale \mbox{of~$\scale^2\ln{\scale}$}.
On the \maxsat\ problem and on the \wmodel\ benchmark, the \opofea\ very significantly outperforms the \opoea.

These results are surprising when considering the nature of FFA -- being invariant under bijective transformations of the objective function, i.e., possessing the strongest invariance property known to us.
FFA never compares objective values directly.
An algorithm applying only FFA would exhibit the same performance on the objective function~\ofel\ as on~\mbox{$g\circ\ofel$}, where~$g$ could be an arbitrary encryption method (which we simulate by setting~$g$ to the \mbox{Md5~checksum} routine in Section~\ref{sec:md5}).

This realization is baffling.
Two central assumptions of black-box optimization are that following a trail of improving objective values tends to be a good idea and that ``nice'' optimization problems should exhibit causality, i.e., small changes to a solution should lead to small changes in its objective value.
Under FFA, neither assumption is used.
As a result, properties such as causality, ruggedness, or deceptiveness of a fitness landscape may have little impact on the algorithm performance.
Interestingly, this does seemingly not necessarily come at a high cost in terms of runtime.
Instead of the cost of the invariance, the limitation of the method seems to be that it requires objective functions that can be discretized and do not take on too many different values.

We finally showed that FFA can be combined with ``normal'' optimization and plugged into more complex algorithms.
We inserted it into the selection step of a memetic algorithm whose local search proceeds without FFA and works directly on the objective values.
Here, an FFA variant purely works as population diversity enhancement mechanism and can improve the result quality that the algorithm produces on the \jssp\ within a budget of five minutes.
Notably, while this algorithm does not belong to the state-of-the-art on the JSSP, it seems to be relatively close to it.
Together with our results on the \maxsat\ problem, this means that FFA might even be helpful in cases bordering to practical relevance.

There are several interesting avenues for future work.
First, we want to also plug FFA into other EAs, such as those in~\cite{CPD2017TAMPARAOEA}.
Second, a theoretical analysis of the properties of FFA could be both interesting and challenging, also from the perspective of black-box complexity.
Third, using FFA is the only approach known to us that can solve encrypted optimization problems.
This could open new types of applications in operations research, machine learning, and artificial intelligence.
Fourth, on problems FFA leads to a slowdown. The question whether this slowdown is proportional to the problem dimension or to the number of possible different objective values deserves an investigation.

Finally, it may be possible to adapt ideas from the research on multi-armed bandits to implement an FFA-like approach:
We envisage an Upper Confidence Bound~\cite{ACBF2002FTAOTMBP}-like algorithm, where one solution per encountered objective value is preserved and treated as bandit arm.
Playing an arm would mean to use the solution as input to mutation and the reward could be~1 if the offspring has a new objective value.%
\section*{Acknowledgments}%
{\small{%
We acknowledge support by %
the National Natural Science Foundation of China under Grant~61673359, %
the Hefei Specially Recruited Foreign Expert Program, %
the Youth Project of the Anhui Natural Science Foundation~1908085QF285, %
the University Natural Science Research Project of Anhui~KJ2019A0835,  %
the Key Research Plan of Anhui~201904d07020002, %
the Major Project of the Research and Development Fund of Hefei University~19ZR05ZDA, %
the Talent Research Fund of Hefei University \mbox{18-19RC26}, %
and %
the \mbox{DIM-RSFI N\textsuperscript{o}~2019-10 C19/1526} project. %
The first author especially wants to thank Dr.\ Carola Doerr for the chance to participate in the Dagstuhl Seminar~19431 with the topic \emph{Theory of Randomized Optimization Heuristics}, which gave us the inspiration to investigate FFA on some basic scenarios where it may even be theoretically tractable. %
\ifCLASSOPTIONcaptionsoff%
\newpage%
\fi%
\bibliographystyle{IEEEtran}%
\small{%

}%
}}%
\newpage%
\begin{IEEEbiography}[{\includegraphics[width=1in,height=1.25in,keepaspectratio]{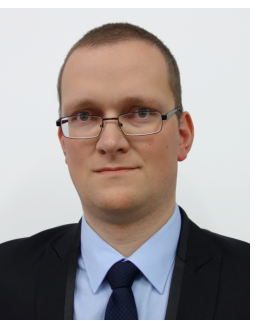}}]{Thomas Weise} obtained his Ph.D.\ in Computer Science from the University of Kassel (Kassel, Germany) in 2009.
He then joined the University of Science and Technology of China (Hefei, Anhui, China), first as PostDoc, later as Associate Professor.
In 2016, he moved to Hefei University (in Hefei) as Full Professor to found the Institute of Applied Optimization (IAO) at the School of Artificial Intelligence and Big Data.
His research interests include metaheuristic optimization and algorithm benchmarking.
He was awarded the title \emph{Hefei Specially Recruited Foreign Expert} of the city Hefei in Anhui, China in 2019.%
\end{IEEEbiography}%
\begin{IEEEbiography}[{\includegraphics[width=1in,height=1.25in,keepaspectratio]{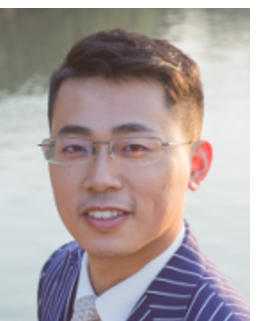}}]{Zhize Wu} obtained his Bachelor from the Anhui Normal University (Wuhu, Anhui, China) in 2012 and his Ph.D.\ in Computer Science from the University of Science and Technology of China (Hefei, Anhui, China) in 2017.
He then joined the Institute of Applied Optimization (IAO) of the School of Artificial Intelligence and Big Data at Hefei University as Lecturer.
His research interest include  image processing, neural networks, deep learning, machine learning in general, and algorithm benchmarking.%
\end{IEEEbiography}%
\begin{IEEEbiography}[{\includegraphics[width=1in,height=1.25in,clip,keepaspectratio]{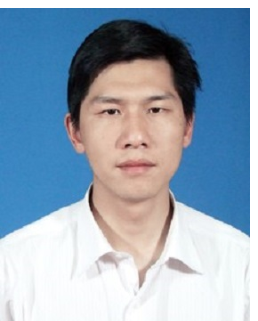}}]{Xinlu Li} received his Bachelor from the Fuyang Normal University (Fuyang, Anhui, China) in 2002, his M.Sc.~degree in Computer Science from Anhui University (Hefei, Anhui, China) in 2009, and his Ph.D.~degree in Computer Science from the TU~Dublin (Dublin, Ireland) in 2019.
His career at Hefei University started as Lecturer in 2009, Senior Lecturer in 2012, in 2018 he joined the Institute of Applied Optimization (IAO), and in 2019 he became Associate Researcher.
From 2014 to 2018, he was Teaching Assistant in the School of Computing of TU~Dublin.
His research interest include Swarm Intelligence and optimization, algorithms which he applies to energy efficient routing protocol design for large-scale Wireless Sensor Networks.%
\end{IEEEbiography}%
\begin{IEEEbiography}[{\includegraphics[width=1in,height=1.25in,keepaspectratio]{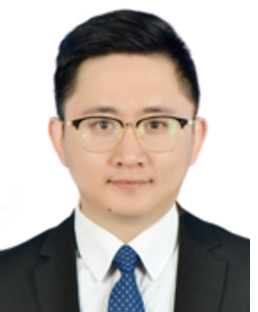}}]{Yan Chen} received his Bachelor from the Anhui University of Science and Technology (Huai'nan, Anhui, China) in 2011, his M.Sc.~degree from the China University of Mining and Technology (Xuzhou, Jiangsu, China) in 2014, and his doctorate from the TU Dortmund (Dortmund, Germany) in 2019.
There, he studied spatial information management and modeling as a member of the Spatial Information Management and Modelling Department of the School of Spatial Planning.
He joined the Institute of Applied Optimization (IAO) of the School of Artificial Intelligence and Big Data at Hefei University as Lecturer in 2019.
His research interests include hydrological modeling, Geographic Information System, Remote Sensing, and Sponge City/Smart City applications, as well as deep learning and optimization algorithms.%
\end{IEEEbiography}%
\setcounter{footnote}{0}%
\let\oldtheffotnote\thefootnote%
\let\thefootnote\relax\footnote{\copyright\ 2020 IEEE. Personal use of this material is permitted. Permission from IEEE must be obtained for all other uses, in any current or future media, including reprinting/republishing this material for advertising or promotional purposes, creating new collective works, for resale or redistribution to servers or lists, or reuse of any copyrighted component of this work in other works.}%
\let\thefootnote\oldtheffotnote%
\setcounter{footnote}{0}%
\end{document}